%% file: main.tex
\documentclass{article}

\usepackage[final]{neurips_2020}

\usepackage[utf8]{inputenc} % allow utf-8 input
\usepackage[T1]{fontenc}    % use 8-bit T1 fonts
\usepackage{hyperref}       % hyperlinks
\usepackage{url}            % simple URL typesetting
\usepackage{booktabs}       % professional-quality tables
\usepackage{amsfonts}       % blackboard math symbols
\usepackage{nicefrac}       % compact symbols for 1/2, etc.
\usepackage{microtype}      % microtypography
\usepackage{appendix} %appendix ToC
\usepackage{pifont}% http://ctan.org/pkg/pifont
\newcommand{\cmark}{\ding{51}}%
\newcommand{\xmark}{\ding{55}}%

\input{macro.tex}

\setcitestyle{square,numbers,comma,sort&compress}

\title{\input{title.tex}}

\author{%
Miles Cranmer$^{1}$ \And
Alvaro Sanchez-Gonzalez$^{2}$ \And
Peter Battaglia$^{2}$ \And
Rui Xu$^{1}$ \AND
Kyle Cranmer$^{3}$ \And
David Spergel$^{4, 1}$ \And
Shirley Ho$^{4,3,1,5}$
\AND
\\
\small $^1$ Princeton University, Princeton, USA\hskip2em
\small $^2$ DeepMind, London, UK\hskip2em \\
\small $^3$ New York University, New York City, USA\hskip2em
\small $^4$ Flatiron Institute, New York City, USA\hskip2em \\
\small $^5$ Carnegie Mellon University, Pittsburgh, USA\hskip2em
}

\begin{document}

\maketitle
\input{content.tex}

\end{document}

%% file: macro.tex
\usepackage{subcaption}
\usepackage{enumitem}
\usepackage{graphicx}
\usepackage{commath}
\usepackage{amsmath}
\usepackage{amssymb}
\usepackage{amsfonts}
\usepackage{mathrsfs}
\usepackage{mathtools}
\usepackage{booktabs}
\usepackage{cleveref}
\usepackage{comment}
\usepackage{multirow}
\usepackage{physics}
\usepackage{color, colortbl, xcolor}
\usepackage{textcomp}
\usepackage{xspace}
\usepackage{ifthen}

\definecolor{Gray}{gray}{0.9}

\newcounter{showComments}
\newboolean{arxiv}   
\setboolean{arxiv}{true} %Arxiv version.
\ifnum \value{showComments}=1
\newcommand{\todo}[1]{{\color{red}\textbf{TODO: #1}}}
\newcommand{\miles}[1]{{\color{purple}\textbf{Miles: #1}}}
\newcommand{\pete}[1]{{\color{red}\textbf{Pete:} #1}}
\newcommand{\shirley}[1]{{\color{green}\textbf{shirley:} #1}}
\newcommand{\alvaro}[1]{{{\color{brown}{[Alvaro: #1]}}}}%
\newcommand{\jess}[1]{{{\color{orange}{[Jess: #1]}}}}%blue is the same color as the citations, so I changed her color to green
\newcommand{\david}[1]{{{\color{orange}{[David: #1]}}}}
\newcommand{\kyle}[1]{{{\color{orange}{[Kyle: #1]}}}}

\else
\newcommand{\todo}[1]{}
\newcommand{\miles}[1]{}
\newcommand{\pete}[1]{}
\newcommand{\shirley}[1]{}
\newcommand{\alvaro}[1]{}
\newcommand{\jess}[1]{}
\newcommand{\david}[1]{}
\newcommand{\kyle}[1]{}
\fi

\newcommand\eureqa{\textit{eureqa}\xspace}

\renewcommand{\v}{\mathbf}

\newcommand\blfootnote[1]{%
  \begingroup
  \renewcommand\thefootnote{}\footnote{#1}%
  \addtocounter{footnote}{-1}%
  \endgroup
}

\renewcommand{\H}{\mathcal{H}}

\newcommand{\Hs}{\mathcal{H}_\text{self}}
\newcommand{\Hp}{\mathcal{H}_\text{pair}}

\renewcommand{\citet}{\cite}

\renewcommand{\citep}{\cite}

%% file: title.tex
Discovering Symbolic Models from Deep Learning with Inductive Biases 

%% file: content.tex
\newcommand\repo{https://github.com/MilesCranmer/symbolic_deep_learning}
\newcommand\video{https://github.com/MilesCranmer/symbolic_deep_learning/blob/master/video_link.txt}
\newcommand\pysr{\textit{PySR}\xspace}
 
\begin{abstract}
    We develop a general approach to distill symbolic representations of a learned deep model by introducing strong inductive biases. We focus on Graph Neural Networks (GNNs). The technique works as follows: we first encourage sparse latent representations when we train a GNN in a supervised setting, then we apply symbolic regression to components of the learned model to extract explicit physical relations. We find the correct known equations, including force laws and Hamiltonians, can be extracted from the neural network. We then apply our method to a non-trivial cosmology example---a detailed dark matter simulation---and discover a new analytic formula which can predict the concentration of dark matter from the mass distribution of nearby cosmic structures. The symbolic expressions extracted from the GNN using our technique also generalized to out-of-distribution-data better than the GNN itself. Our approach offers alternative directions for interpreting neural networks and discovering novel physical principles from the representations they learn.
 \blfootnote{Code for our models and experiments can be found at \url{\repo}.}
\end{abstract}

\section{Introduction}
 
{\it The miracle of the appropriateness of the language of mathematics for the formulation of the laws of physics is a wonderful gift which we neither understand nor deserve. We should be grateful for it and hope that it will remain valid in future research and that it will extend, for better or for worse, to our pleasure, even though perhaps also to our bafflement, to wide branches of learning.}—Eugene Wigner ``{The Unreasonable Effectiveness of Mathematics in the Natural Sciences}'' \citet{wigner}.
 
For thousands of years, science has leveraged models made out of closed-form symbolic expressions, thanks to their many advantages: algebraic expressions are usually compact, present explicit interpretations, and generalize well.
However, finding these algebraic expressions is difficult.
Symbolic regression is one option: a supervised machine learning technique that assembles analytic functions to model a given dataset. 
However, typically one uses genetic algorithms---essentially a brute force procedure as in \cite{schmidt2009distilling}---which scale exponentially with the number of input variables and operators. Many machine learning problems are thus intractable for traditional symbolic regression.
 
On the other hand, deep learning methods allow efficient training of complex models on high-dimensional datasets. However, these learned models are black boxes, and difficult to interpret. Furthermore, generalization is difficult without prior knowledge about the data imposed directly on the model. Even if we impose strong inductive biases on the models to improve generalization, the learned parts of networks typically are linear piece-wise approximations which extrapolate linearly (if using ReLU as activation \cite{montufar2014number}).

Here, we propose a general framework to leverage the advantages of both deep learning and symbolic regression. As an example, we study Graph Networks (GNs or GNNs) \citep{battaglia2018relational} as they have strong and well-motivated inductive biases that are very well suited to problems we are interested in. Then we apply symbolic regression to fit different internal parts of the learned model that operate on reduced size representations. The symbolic expressions can then be joined together, giving rise to an overall algebraic equation equivalent to the trained GN. Our work is a generalized and extended version of that in \cite{last}.
 
We apply our framework to three problems---rediscovering force laws, rediscovering Hamiltonians, and a real world astrophysical challenge---and demonstrate that we can drastically improve generalization, and distill plausible analytical expressions. We not only recover the injected closed-form physical laws for Newtonian and Hamiltonian examples, we also derive a new interpretable closed-form analytical expression that can be useful in astrophysics.
 
\section{Framework}
 
\begin{figure*}[t]
    \centering
    \includegraphics[width=0.95\textwidth]{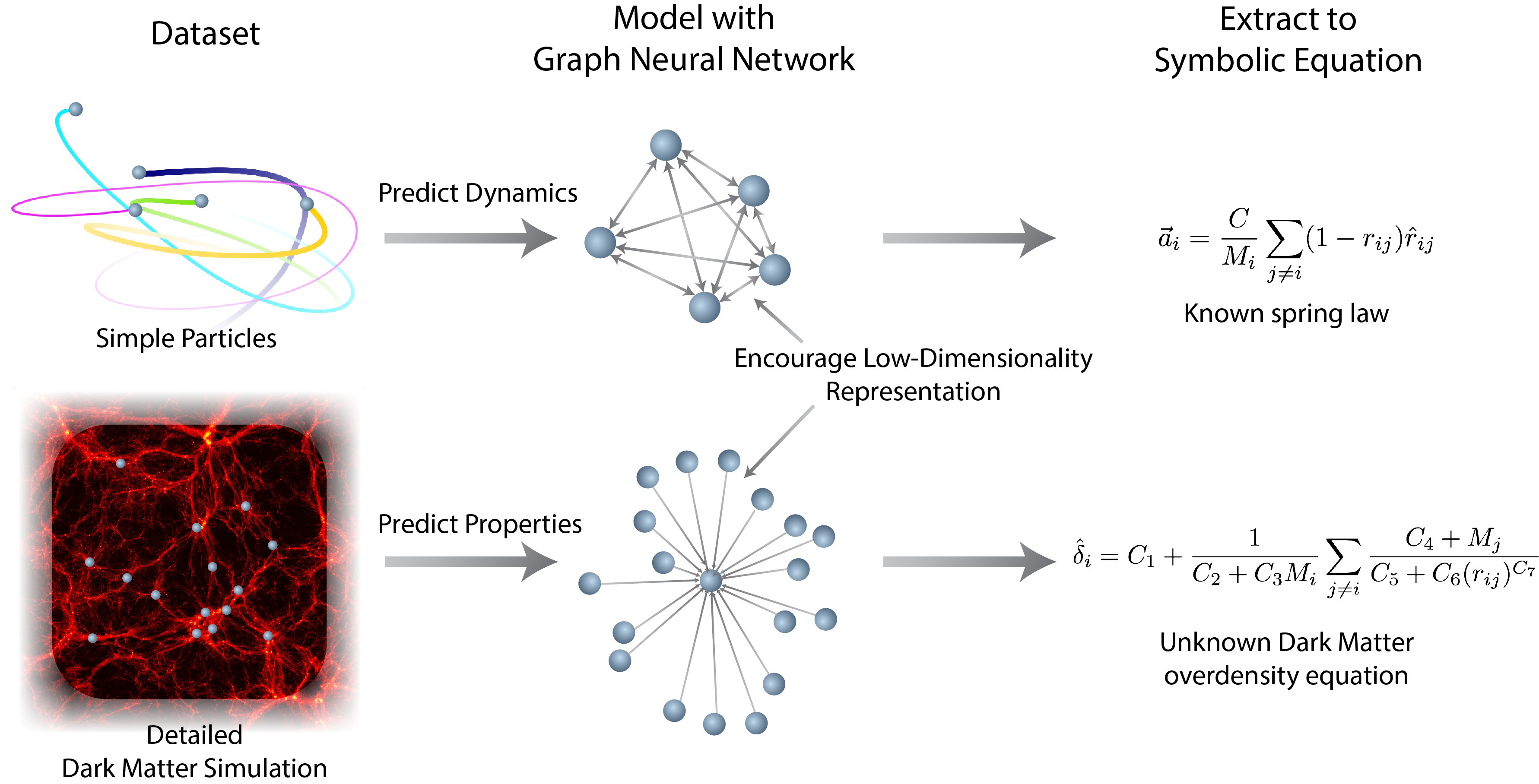}
    \caption{A cartoon depicting how we extract physical equations from a dataset. }
    \label{fig:setup}
\end{figure*}

Our framework can be summarized as follows. (1) Engineer a deep learning model with a separable internal structure that provides an inductive bias well matched to the nature of the data. Specifically, in the case of interacting particles, we use Graph Networks as the core inductive bias in our models. (2) Train the model end-to-end using available data. (3) Fit symbolic expressions to the distinct functions learned by the model internally. (4) Replace these functions in the deep model by the symbolic expressions. This procedure with the potential to discover new symbolic expressions for non-trivial datasets is illustrated in \cref{fig:setup}.
 
\paragraph{Particle systems and Graph Networks.}
In this paper we focus on problems that can be well described as interacting particle systems. Nearly all of the physics we experience
in our day-to-day life can be described in terms of interactions rules
between particles or entities, so this is broadly relevant. Recent work has leveraged the inductive biases of Interaction Networks (INs) \citep{battaglia2016interaction} in their generalized form, the \emph{Graph Network}, a type of Graph Neural Network \citep{scarselli2009graph,bronstein2017geometric,gilmer2017neural},
to learn models of particle systems in many physical domains \citep{battaglia2016interaction,chang2016compositional,sanchez2018graph,mrowca2018flexible,li2018learning,kipf2018neural,bapst2020unveiling,sanchez2020learning}.
 
Therefore we use Graph Networks (GNs) at the core of our models, and incorporate into them physically motivated inductive biases appropriate for each of our case studies. Some other interesting approaches for learning low-dimensional general dynamical models include \citet{packard1980geometry,daniels_automated_2015,jaques_2019}. Other related work which studies the physical reasoning abilities of deep models include \cite{physreasoning,phyre,reasonabout,glass}.
 
\begin{figure*}[h]
    \centering
    \includegraphics[width=0.8\textwidth]{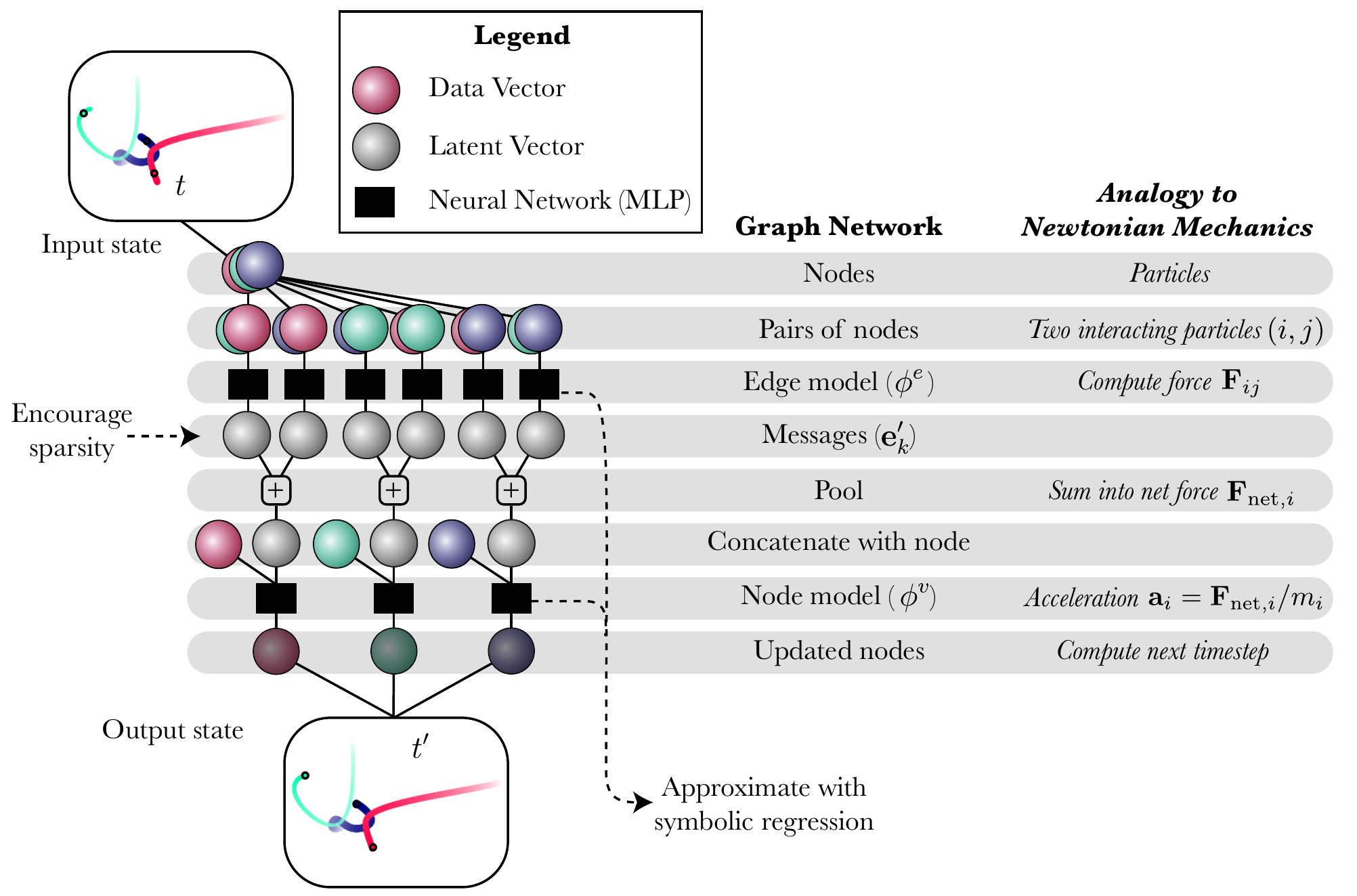}
    \caption{An illustration of the internal structure of the graph neural network
    we use in some of our experiments.
    Note that the comparison to Newtonian mechanics is purely for explanatory purposes, but is not explicit.
    Differences include: the ``forces'' (messages) are often high dimensional,
    the nodes need not be physical particles, $\phi^e$ and $\phi^v$ are arbitrary learned functions,
    and the output need not be an updated state.
    However, the rough equivalency between this architecture and physical
    frameworks allows us to interpret learned formulas in terms of existing
    physics.
    }%
    \label{fig:gnstruct}
\end{figure*}

Internally, GNs are structured into three distinct components: an edge model, a node model, and a global model, which take on different but explicit roles in a regression problem. The edge model, or ``message function,'' which we denote by $\phi^e$, maps from a pair of nodes ($\mathbf{v}_i, \mathbf{v}_j \in \mathbb{R}^{L_v}$)  connected by an edge in a graph together with some vector information for the edge, to a message vector. These message vectors are summed element-wise for each receiving node over all of their sending nodes, and the summed vector is passed to the node model. The node model, $\phi^v$, takes the receiving node and the summed message vector, and computes an updated node: a vector representing some property or dynamical update. Finally, a global model $\phi^u$ aggregates all messages and all updated nodes and computes a global property. $\phi^e$, $\phi^v$, $\phi^u$ are usually approximated using multilayer-perceptrons, making them differentiable end-to-end. More details on GNs are given in the appendix. We illustrate the internal structure of a GN in \cref{fig:gnstruct}.
 
GNs are the ideal candidate for our approach due to their inductive biases shared by many physics problems. (a) They are equivariant under particle permutations. (b) They are differentiable end-to-end and can be trained efficiently using gradient descent. (c) They make use of three separate and interpretable internal functions $\phi^e$, $\phi^v$, $\phi^u$, which are our targets for the symbolic regression. 
GNs can also be embedded with additional symmetries as in \citet{bekkers, liegnn}, but we do not implement these.
 
\paragraph{Symbolic regression.}
After training the Graph Networks, we use the symbolic regression package \eureqa \citep{schmidt2009distilling} to perform symbolic regression and fit compact closed-form analytical expressions to $\phi^e$, $\phi^v$, and $\phi^u$ independently. \eureqa works by using a genetic algorithm to combine algebraic expressions stochastically. The technique is similar to natural selection, where the ``fitness'' of each expression is defined in terms of simplicity and accuracy. The operators considered in the fitting process are $+, -, \times, /, >, <,$ \textasciicircum$, \exp, \log, \textsc{IF}(\cdot,\cdot,\cdot)$ as well as real constants. After fitting expressions to each part of the graph network, we substitute the expressions into the model to create an alternative analytic model. We then refit any parameters in the symbolic model to the data a second time, to avoid the accumulated approximation error. Further details are given in the appendix.

This approach gives us an explicit way of interpreting the trained weights of a structured neural network. An alternative way of interpreting GNNs is given in \citet{explainer}.
This new technique also allows us to extend symbolic regression to high-dimensional datasets,
where it is otherwise intractable.
As an example,
consider attempting to discover the relationship
between a scalar and a time series,
given data $\{(z_i, \{\vb x_{i,1}, \vb x_{i,2}, \ldots  \vb x_{i,100}\} \}$,
where $z_i\in \mathbb{R}$ and
$\vb x_{i,j}\in \mathbb{R}^5$.
Assume the true relationship as
$z_i =y_i^2$, for $y_i = \sum_{j=1}^{100} y_{i,j},$
${y_{i,j} = \exp(x_{i,j,3}) + \cos(2 x_{i, j, 1})}.$
Now, in a learnable model, assume an inductive bias
${z_i = f(\sum_{j=1}^{100} g(\vb x_{i, j}))}$
for scalar functions $f$ and $g$.
If we need to consider $10^9$ equations for both
$f$ and $g$,
then a standard symbolic regression search
would need to consider their combination,
leading to $(10^9)^2=10^{18}$
equations in total.
But if we first fit a neural network for $f$ and $g$,
and after training, fit an equation to $f$ and $g$ \textit{separately},
we only need to consider $2\times 10^{9}$ equations.
In effect, we factorize high-dimensional
datasets into smaller sub-problems that are tractable
for symbolic regression.

We emphasize that this method is not a new symbolic regression
technique by itself; rather, it is a way of extending
any existing symbolic regression method to high-dimensional
datasets by the use of a neural network with a
well-motivated inductive bias.
While we chose \eureqa for our experiments based on
its efficiency and ease-of-use, we could have chosen
another low-dimensional symbolic regression package, such as 
our new high-performance package
\textit{PySR}\footnote{\url{https://github.com/MilesCranmer/PySR}} \citet{pysr}.
Other community
packages such as \citet{dcgp,eql,eql2,fortin,udrescu,pge,guimera,grammarvae,lili},
could likely also be used
and achieve similar results (although
\citet{udrescu} 
is unable to fit the constants required for the tasks here).
Ref.~\citet{eql} is an interesting approach that uses 
gradient descent on a pre-defined equation up to some depth,
parametrized with a neural network,
instead of genetic algorithms;
\citet{grammarvae} uses gradient descent on a latent embedding
of an equation; and \citet{lili} demonstrates Monte Carlo
Tree Search as a symbolic regression technique, using
an asymptotic constraint as input to a neural network
which guides the search.
These could all be used as drop-in replacements for \eureqa
here to extend their algorithms to high-dimensional datasets.

We also note several exciting packages for symbolic regression
of partial differential equations on gridded data:
\citet{deepymod,brunton,atkinson,universalode,pidl,fluidpdealg}.
These either use sparse regression
of coefficients over a library of PDE terms,
or a genetic algorithm.
While not applicable to our
use-cases, these would be interesting
to consider for future extensions to gridded PDE data.

\paragraph{Compact internal representations.}
While training, we encourage the model to use compact internal representations for latent hidden features (e.g., messages) by adding regularization terms to the loss (we investigate using L$_1$ and KL penalty terms with a fixed prior, see more details in the Appendix). One motivation for doing this is based on \emph{Occam's Razor}: science always prefers the simpler model or representation of two which give similar accuracy. Another stronger motivation is that if there is a law that perfectly describes a system in terms of summed message vectors in a compact space (what we call a linear latent space), then we expect that a trained GN, with message vectors of the same dimension as that latent space, will be mathematical rotations of the true vectors. We give a mathematical explanation of this reasoning in the appendix, and emphasize that while it may seem obvious now, our work is the first to demonstrate it. More practically, by reducing the size of the latent representations, we can filter out all low-variance latent features without compromising the accuracy of the model, and vastly reducing the dimensionality of the hidden vectors. This makes the symbolic regression of the internal models more tractable.
 
\paragraph{Implementation details.} We write our models with PyTorch \citep{torch} and PyTorch Geometric\citep{geom}. We train them with a decaying learning schedule using Adam \citep{kingma}. The symbolic regression technique is described in \cref{sec:force}. More details are provided in the Appendix.
 
\section{Case studies}
\label{sec:case}
 
In this section we present three specific case studies where we apply our proposed framework using additional inductive biases.
 
\paragraph{Newtonian dynamics.}
Newtonian dynamics describes the dynamics of particles according to Newton's law of motion: the motion of each particle is modeled using incident forces from nearby particles, which change its position, velocity and acceleration.  Many important forces in physics (e.g., gravitational force $-\frac{G m_1 m_2}{r^2} \hat{r}$) are defined on pairs of particles, analogous to the message function $\phi^e$ of our Graph Networks. The summation that aggregates messages is analogous to the calculation of the net force on a receiving particle. Finally, the node function, $\phi^v$, acts like Newton's law: acceleration equals the net force (the summed message) divided by the mass of the receiving particle.
 
To train a model on Newtonian dynamics data, we train the GN to predict the instantaneous acceleration of the particle against that calculated in the simulation.
While Newtonian mechanics inspired the original development of INs, never before
has an attempt to distill the relationship between the forces
and the learned messages been successful. 
When applying the framework to this Newtonian dynamics problem (as illustrated in \cref{fig:setup}), we expect the model trained with our framework to discover that the optimal dimensionality of messages should match the number of spatial dimensions. We also expect to recover algebraic formulas for pairwise interactions, and generalize better than purely learned models. We refer our readers to \cref{sec:force} and the Appendix for more details.

\paragraph{Hamiltonian dynamics.}
 
Hamiltonian dynamics describes a system's
total energy $\H(\vb{q}, \vb{p})$ as a function of
its canonical coordinates $\mathbf{q}$ and momenta $\mathbf{p}$---e.g.,
each particle's position and momentum.
The dynamics of the system change
perpendicularly to the gradient of $\H$:
$\dv{\vb{q}}{t} = \pdv{\H}{\vb{p}}, \dv{\vb{p}}{t} = -\dv{\H}{\vb{q}}.$%\frac{\partial{\H}}{\partial{\mathbf{q}}}.$

Here, we will use a variant of a
Hamiltonian Graph Network (HGN) \cite{sanchez} to learn $\H$
for the Newtonian dynamics data.
This model is a combination
of a Hamiltonian Neural Network \citet{greydanus,toth} and GN.
In this case, the global model $\phi^u$ of the GN will output a
single scalar value for the entire system representing the energy, and hence the GN will have the same functional form as a Hamiltonian. 
By then taking the partial derivatives of the GN-predicted $\H$ with respect to the position and momentum, $\mathbf{q}$ and $\mathbf{p}$, respectively, of the input nodes, we will be able to calculate the updates to the momentum and position.
We impose a modification to the HGN to facilitate its interpretability,
and name this the ``Flattened HGN'' or FlatHGN:
instead of summing high-dimensional encodings of nodes to
calculate $\phi^u$, we instead set it to be a sum of scalar
pairwise interaction terms, $\Hp$
and a per-particle term, $\Hs$. This is because many physical
systems can be exactly described this way.
This is a Hamiltonian version of the Lagrangian Graph Network in \cite{lagrangian},
and is similar to \cite{wang2019}.
This is still general enough to express many physical systems,
as nearly all of physics can be written as summed interaction energies,
but could also be relaxed in the context of the framework.
 
Even though the model is trained end-to-end, we expect our framework to allow us to extract analytical expressions for both the per-particle kinetic energy, and the scalar pairwise potential energy. 
We refer our readers to our \cref{sec:hamiltonian} and the Appendix for more details.

\paragraph{Dark matter halos for cosmology.}
 
We also apply our framework to a dataset generated from state-of-the-art dark matter simulations \citep{quijote}. We predict a property (``overdensity'') of a dark matter blob (called a ``halo'') from the properties (positions, velocities, masses) of halos nearby. We would like to extract this relationship as an analytic expression so we may interpret it theoretically. This problem differs from the previous two use cases in many ways, including (1) it is a real-world problem where an exact analytical expression is unknown; (2) the problem does not involve dynamics, rather, it is a regression problem on a static dataset; and (3) the dataset is not made of particles, but rather a grid of density that has been grouped and reduced to handmade features. Similarly, we do not know the dimensionality of interactions, should a linear latent space exist. We rely on our inductive bias to find the optimal dimensional of the problem, and then yield an interpretable model that performs better than existing analytical approximations. We refer our readers to our \cref{sec:cosmo} and the Appendix for further details. 
 
\section{Experiments \& results}
\label{sec:exp}

\begin{figure*}[h]
    \centering
    \href{\video}{\includegraphics[width=\textwidth]{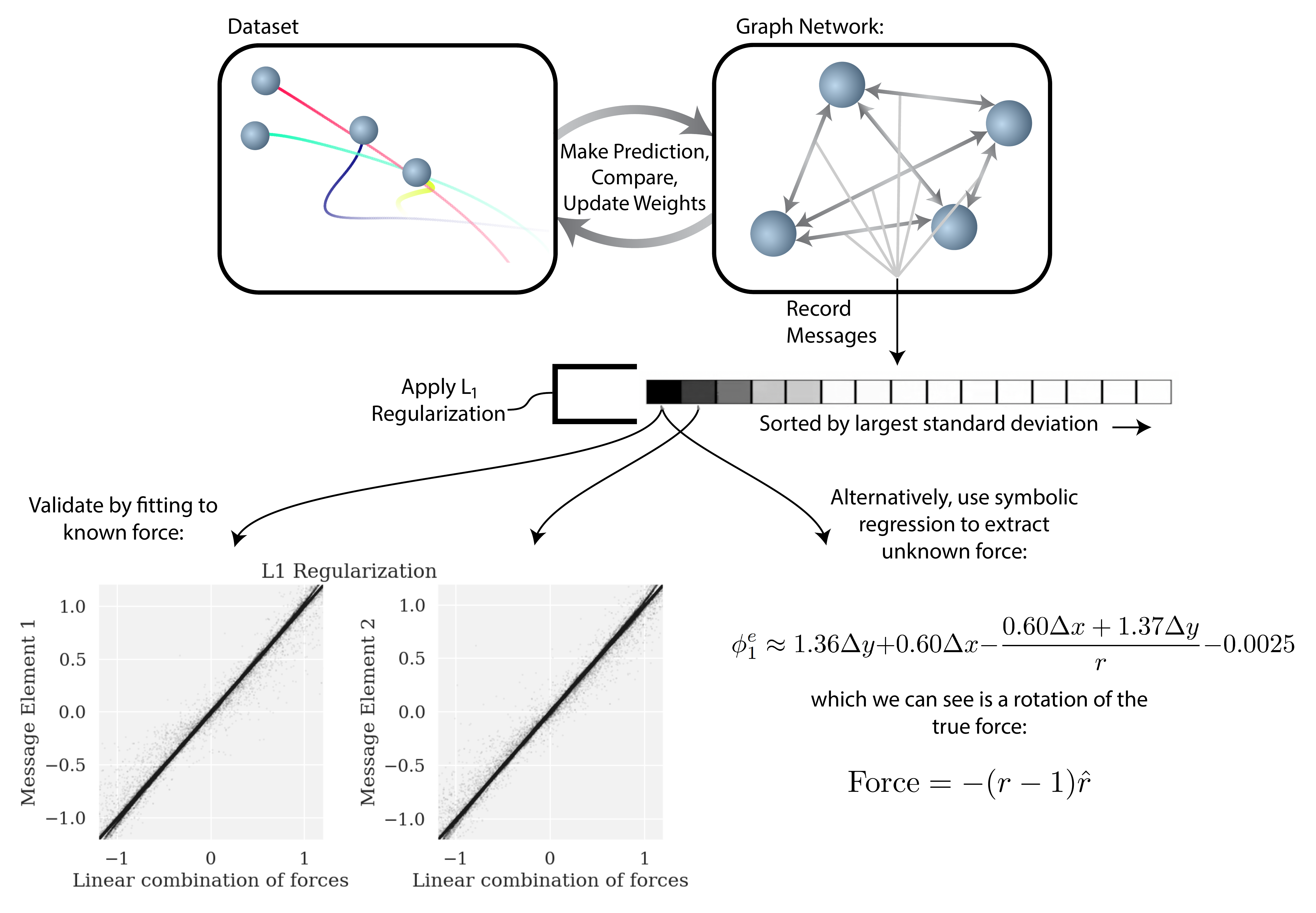}}
    \caption{A diagram showing how we implement and exploit
    our inductive bias on GNs. A video of this figure during training
    can be seen by going to the URL \url{\video}.}%
    \label{fig:video}
\end{figure*}
 
\subsection{Newtonian dynamics}
\label{sec:force}

We train our Newtonian dynamics GNs on data for simple N-body systems with known force laws. We then apply our technique to recover the known force laws via the representations learned by the message function $\phi^e$.

\paragraph{Data.}
The dataset consists of N-body particle simulations in two and three dimensions, under different interaction laws.
We used the following forces:
(a) $1/r$ orbital force: $-m_1 m_2\hat{r}/r $;
(b) $1/r^2$ orbital force $-m_1 m_2\hat{r}/r^2$;
(c) charged particles force $q_1 q_2\hat{r}/r^2$;
(d) damped springs with $\abs{r-1}^2$ potential
and damping proportional and opposite to speed;
(e) discontinuous forces, $-\{0, r^2\}\hat{r}$, switching to $0$ force for $r < 2$; and
(f) springs between all particles, a $(r-1)^2$ potential.
The simulations themselves contain masses and charges of 4 or 8 particles, with positions, velocities, and accelerations as a function of time.
Further details of these systems
are given in the appendix, with example trajectories
shown in \cref{fig:simulations}. 
 
\paragraph{Model training.}
The models are trained to predict instantaneous acceleration for every particle
given the current state of the system. To investigate the importance of the size of the message representations for interpreting the messages
as forces, we train our GN using 4 different strategies:
1. {Standard}, a GN with 100 message components;
2. {Bottleneck}, a GN with the number of message components matching the dimensionality of the problem (2 or 3);
3. L$_1$, same as ``Standard'' but using a L$_1$ regularization loss term on the messages with a weight of $10^{-2}$; and 
4. {KL} same as ``Standard'' but regularizing the messages using the Kullback-Leibler (KL) divergence with respect to Gaussian prior.
Both the L$_1$ and {KL} strategies encourage
the network to find compact representations for the message vectors, using
different regularizations.
We optimize the mean absolute loss
between the predicted acceleration and the true acceleration of each node.
Additional training details are given in the appendix and found in the codebase. 
 
\paragraph{Performance comparison.}
To evaluate the learned models, we generate a new dataset from a different random seed. We find that the model with L$_1$ regularization has the greatest prediction performance in most cases (see \cref{tbl:losses}).
It is worth noting that the bottleneck model,
even though it has the correct dimensionalty,
performs worse than the model using L$_1$ regularization
under limited training time. 
We speculate that this may connect to the
lottery ticket hypothesis \citet{lottery}.

\paragraph{Interpreting the message components.}
As a first attempt to interpret the information in the message components, we pick the $D$ message features (where $D$ is the dimensionality of the simulation) with the highest variance (or KL divergence), and fit each to a linear combination of the true force components. We find that while the GN trained in the {Standard} setting does not show strong correlations with force components (also seen in \cref{fig:fitted}), all other models for which the effective message size is constrained explicitly (bottleneck) or implicitly (KL or L$_1$) to be low dimensional yield messages that are highly correlated with the true forces (see \cref{tbl:forcefit} which indicates the fit errors with respect to the true forces), with the model trained with L$_1$ regularization showing the highest correlations. An explicit demonstration that the messages in a graph network learn forces has not been observed before our work.
 
The messages in these models are thus explicitly interpretable as forces. The video at \url{\video} (\cref{fig:video}) shows a fit of the message components over time during training, showing how the model discovers a message representation that is highly correlated with a rotation of the true force vector in an unsupervised way. 
 
\begin{table}[h]
    \centering
\begin{tabular}{@{}lcccc@{}}
    \toprule
        Sim. &         Standard &                Bottleneck &                   L$_1$       &                   KL      \\ \midrule
    Charge-2 &           0.016  &                   0.947   &                   0.004       &                 0.185     \\                                      
    Charge-3 &           0.013  &                   0.980   &                   0.002       &                 0.425     \\                             
    $r^{-1}$-2 &         0.000  &                   1.000   &                   1.000       &                 0.796     \\                                 
    $r^{-1}$-3 &         0.000  &                   1.000   &                   1.000       &                 0.332     \\                                
    $r^{-2}$-2 &         0.004  &                   0.993   &                   0.990       &                 0.770     \\                                
    $r^{-2}$-3 &         0.002  &                   0.994   &                   0.977       &                 0.214     \\                                
    Spring-2 &           0.032  &                   1.000   &                   1.000       &                 0.883     \\                                
    Spring-3 &           0.036  &                   0.995   &                   1.000       &                 0.214     \\ \bottomrule\\
\end{tabular}
\caption{The $R^2$ value of a fit of
    a linear combination of true force components to the message components
    for a given model (see text).
    Numbers close to 1 indicate the messages and true force are strongly correlated.
    Successes/failures of force law symbolic regression are tabled in the appendix.
    }
\label{tbl:forcefit}
\end{table}
 
\paragraph{Symbolic regression on the internal functions.}
We now demonstrate symbolic regression to extract
force laws from the messages, without using prior knowledge for each force's form. To do this, we record the most significant message component of $\phi^e$, which we refer to as $\phi^e_1$, over random samples of the training dataset. The inputs to the regression are
$m_1, m_2, q_1, q_2, x_1, x_2, \ldots$ (mass, charge, x-position
of receiving and sending node) as well as simplified variables to help the symbolic regression: e.g., $\Delta x$
for $x$ displacement, and $r$ for distance.
 
We then use \eureqa to fit the $\phi^e_1$ to the inputs by minimizing the mean absolute error (MAE) over various analytic functions. Analogous to Occam's razor, we find the ``best'' algebraic model by asking \eureqa to provide multiple candidate fits at different complexity levels (where complexity is scored as a function of the number and the type of operators, constants and input variables used), and select the fit that maximizes the fractional drop in mean absolute error (MAE) over the increase in complexity from the next best model: $(-\Delta \log(\text{MAE}_c)/\Delta c)$.
From this, we recover many analytical expressions (this is tabled in the appendix)
that are equivalent to the simulated force laws ($a, b$ indicate learned constants):
\begin{itemize}
    \item Spring, 2D, L$_1$ (expect $\phi^e_1\approx (\v{a}\cdot (\Delta x, \Delta y)) (r - 1) + b$). 
$$\phi^e_1\approx 1.36 \Delta y + 0.60 \Delta x - \frac{0.60 \Delta x + 1.37 \Delta y}{r} - 0.0025 $$
    \item $1/r^2$, 3D, Bottleneck (expect $\phi^e_1\approx \frac{\v{a}\cdot (\Delta x, \Delta y, \Delta z)}{r^3} + b$). 
$$\phi^e_1\approx \frac{0.021\Delta x m_2 - 0.077 \Delta y m_2}{r^3}$$
    \item Discontinuous, 2D, L$_1$ (expect $\phi^e_1\approx \textsc{IF}(r>2, (\v{a}\cdot (\Delta x, \Delta y, \Delta z)) r, 0) + b$). 
            $$\phi^e_1 \approx \textsc{IF}(r>2,
            0.15 r \Delta y + 0.19 r \Delta x,
            0) - 0.038$$
\end{itemize}
 
Note that reconstruction does not always succeed, especially
for training strategies other than L$_1$ or {bottleneck} models that cannot successfully find compact representations of the right dimensionality (see some examples in Appendix).
 
\subsection{Hamiltonian dynamics}
\label{sec:hamiltonian}
 
Using the same datasets from the Newtonian dynamics case study, we also train our ``FlatHGN,'' with the Hamiltonian inductive bias, and demonstrate that we can extract scalar potential energies, rather than forces, for all of our problems. For example, in the case of charged particles, with expected potential ($\Hp \approx \frac{a q_1q_2}{r}$), symbolic regression applied to the learned message function yields\footnote{We have removed constant terms that don't depend on the position or momentum as those are just arbitrary offsets in the Hamiltonian which don't have an impact on the dynamics. See Appendix for more details.}: ${\Hp\approx \frac{0.0019 q_1 q_2}{r}.}$

It is also possible to fit the per-particle term $\Hs$, however, in this case the same kinetic energy expression is recovered for all systems.
In terms of performance results, the Hamiltonian models are comparable to that of the L$_1$ regularized model across all datasets (See Supplementary results table).
 
Note that in this case, by design, the ``FlatHGN`` has a message function with a dimensionality of 1 to match the output of the Hamiltonian function which is a scalar, so no regularization is needed, as the message size is directly constrained to the right dimension. 
 
\subsection{Dark matter halos for cosmology}
\label{sec:cosmo}
 
Now, one may ask: ``will this strategy also work for
general regression problems,
non-trivial datasets, complex interactions,
and unknown laws?'' Here we give an example
that satisfies all four of these concerns, using
data from a gravitational simulation of the Universe.
 
Cosmology studies the evolution of the Universe
from the Big Bang to the complex galaxies and stars we see
today \citet{cosmotext}. The interactions of various types of matter and energy drive this evolution.
Dark Matter alone consists of $\approx$ 85\% of the total matter in the Universe \cite{spergel,planck}, and therefore is extremely important for the development of galaxies. Dark matter particles clump together and act as gravitational basins called ``halos'' which pull regular baryonic matter together to produce stars, and form larger structures such as filaments and galaxies.
It is an important question in cosmology to predict properties of dark matter halos based on their ``environment,'' which consist of other nearby dark matter halos. 
Here we study the following problem: how can we predict the excess amount of matter (in comparison to its surroundings, $\delta = \frac{\rho - \ev{\rho}}{\ev{\rho}}$) for a dark matter halo based on its properties and those of its neighboring dark matter halos?
 
A hand-designed estimator for the functional form of
$\delta_i$ for halo $i$
might correlate $\delta$ with the mass of the same
halo, $M_i$, as well as the mass within 20 distance units (we decide to use 20 as the smoothing radius):
$\sum_{j\neq i}^{\abs{\mathbf{r}_i - \mathbf{r}_j} < 20} M_j$. 
The intuition behind this scaling is described in \cite{frenk}.
Can we find a better equation that we can fit better to the data,
using our methodology?

\paragraph{Data, training and symbolic regression.}
We study this problem with the open sourced
N-body dark matter simulations from \cite{quijote}.
We choose the zeroth simulation in this dataset, at the final time step (current day Universe), which contains 215,854 dark matter halos. Each halo has mass $M_i$, position $\mathbf{r}_i$, and velocity $\mathbf{v}_i$. 
We also compute the smoothed overdensity $\delta_i$
at the location of the center of  each halo.
We convert this set of halos into a graph by connecting halos within
fifty distance units (each distance unit is approximately 3 million light years long) of each other. This results
in 30,437,218 directional 
edges between halos, or 71 neighbors per halo on average.
We then attempt to predict $\delta_i$ for each halo with
a GN.
Training details are the same as for the Newtonian simulations,
but we switch to 500 hidden units after hyperparameter tuning based
on GN accuracy.
 
The GN trained with L$_1$ appears to have messages
containing only 1 informative feature, so we extract message samples for this component of the messages
over the training set for random pairs of halos, and node function
samples for random receiving halos and their summed messages.
The formula extracted by the algorithm is given in \cref{tbl:new} as ``Best, with mass''.
The form of the formula is new and it captures a well-known relationship between halo mass and environment: bias-mass relationship. 
We refit the parameters in the formula on the original
training data to avoid accumulated approximation error from the multiple
levels of function fitting. 
We achieve a loss of 0.0882 where the hand-designed formula achieves a loss of 0.121.  It is quite surprising that our formula extracted by our approach is able to achieve a better fit than the formula hand-designed by scientists. 
 
\newcommand{\STAB}[1]{\begin{tabular}{@{}c@{}}#1\end{tabular}}
 
\begin{table*}[]
\centering
\begin{tabular}{@{}lcccc@{}}
\toprule
& Test & Formula & Summed Component & ${\scriptstyle \ev{\abs{\delta_i - \hat{\delta}_i}}}$\\ \hline \hline
\multirow{2}{*}{\STAB{\rotatebox[origin=c]{90}{Old~}}} & Constant & $\hat{\delta}_i = C_1$ & N/A & 0.421
\\ \cline{2-5}
& Simple & $\hat{\delta}_i = C_1 + (C_2 + M_i C_3)e_i$ & $e_i = \sum_{j\neq i}^{\abs{\mathbf{r}_i - \mathbf{r}_j} < 20} M_j$ & 0.121 
\\ \hline
\multirow{2}{*}{\STAB{\rotatebox[origin=c]{90}{New~}}} & Best, without mass & $\hat{\delta}_i = C_1 + \frac{e_i}{C_2 + C_3 e_i \abs{\mathbf{v}_i}}$ & $e_i = \sum_{j\neq i} \frac{C_4 + \abs{\mathbf{v}_i - \mathbf{v}_j}}{C_5 + (C_6\abs{\mathbf{r_i} - \mathbf{r_j}})^{C_7}} $& 0.120 \\ \cline{2-5}
& Best, with mass & $\hat{\delta}_i = C_1 + \frac{e_i}{C_2 + C_3 M_i}$ & $e_i = \sum_{j\neq i} \frac{C_4 + M_j}{C_5 + (C_6 \abs{\mathbf{r}_i - \mathbf{r}_j})^{C_7}}$& 0.0882 \\
\bottomrule
\end{tabular}
\caption{A comparison of both known and discovered formulas
for dark matter overdensity. $C_i$ indicates fitted parameters, which are given in the appendix.}
\label{tbl:new}
\end{table*}
 
The formula makes physical sense.
Halos closer to the dark matter halo of interest should influence its properties more, and thus the summed function scales inversely with distance. Similar to the hand-designed formula, the overdensity should
scale with the total matter density nearby, and we see this in that we
are summing over mass of neighbors.
The other differences are very interesting, and less clear;
we plan to do detailed interpretation of these results
in a future astrophysics study.
 
As a followup, we also calculated if we could predict
the halo overdensity from only velocity and position information.
This is useful because the most direct observational
information available is in terms of halo velocities.
We perform an identical analysis without mass information,
and find a curiously similar formula. The 
relative speed between two neighbors can be seen as a proxy for mass,
which is seen in \cref{tbl:new}.
This makes sense as a more massive object will have
more gravity, accelerating falling particles near it
to faster speeds.
This formula is also new to cosmologists, and can in principle
help push forward cosmological analysis.
 
\paragraph{Symbolic generalization.}
 
As we know that our physical world is well described by mathematics,
we can use it as a powerful prior
for creating new models of our world.
Therefore, if we distill a neural network into a simple algebra,
will the algebra generalize better to unseen data?
Neural nets excel at learning in high-dimensional
spaces, so perhaps, by combining both of these
types of models, one can leverage the unique
advantages of each.
Such an idea is discussed in detail in \cite{marcus}.
 
Here we study this on the cosmology example
by masking 20\% of the data: halos which have $\delta_i > 1$.
We then proceed through the same training procedure as before, 
learning a GN to predict $\delta$ with L$_1$ regularization, 
and then extracting messages for examples in the training set.
Remarkably, we obtain a functionally identical expression when extracting the formula
from the graph network on this subset of the data.
We fit these constants to the same masked
portion of data on which the graph network was trained.
The graph network itself obtains an average error
${\scriptsize \ev*{\abs*{\delta_i - \hat{\delta}_i}}= 0.0634}$
on the training set, and 0.142
on the out-of-distribution data. Meanwhile, the
symbolic expression achieves 0.0811 on the training set,
but 0.0892 on the out-of-distribution data.
Therefore, for this problem, it seems a symbolic expression
generalizes much better than the very graph neural network it was extracted from.
This alludes back to Eugene Wigner's article: the
language of simple, symbolic models is remarkably effective
in describing the universe.
 
\section{Conclusion}
We have demonstrated a general approach for imposing
physically motivated inductive biases on GNs
and Hamiltonian GNs to learn interpretable representations,
and potentially improved zero-shot generalization.
Through experiment, we have shown that GN models
which implement a bottleneck or L$_1$ regularization in the message
passing layer, or a Hamiltonian GN flattened to pairwise and self-terms,
can learn message representations equivalent to linear transformations
of the true force vector or energy.
We have also demonstrated a generic technique for finding an unknown force law
from these models: symbolic regression is capable of
fitting explicit equations to our trained model's message function.
We repeated this for energies instead of forces
via introduction of the ``Flattened Hamiltonian Graph Network.''
Because GNs have more explicit substructure than their more homogeneous deep learning relatives (e.g., plain MLPs, convolutional networks), we can draw more fine-grained interpretations of their learned representations and computations.
Finally, we have demonstrated our algorithm on a non-trivial dataset,
and discovered a new law for cosmological dark matter.
 
\clearpage
 
\begin{ack}
    Miles Cranmer would like to thank Christina Kreisch and Francisco Villaescusa-Navarro for assistance with the cosmology dataset; and Christina Kreisch, Oliver Philcox, and Edgar Minasyan for comments on a draft of the paper. The authors would like to thank the reviewers for insightful feedback that improved this paper. Shirley Ho and David Spergel’s work is supported by the Simons Foundation.
 
Our code made use of the following Python packages:
\texttt{numpy},
\texttt{scipy},
\texttt{sklearn},
\texttt{jupyter},
\texttt{matplotlib},
\texttt{pandas},
\texttt{torch},
\texttt{tensorflow},
\texttt{jax}, and
\texttt{torch\_geometric}
\cite{numpy,scipy,sklearn,jupyter,matplotlib,pandas,torch,tensorflow,jax,geom}.
\end{ack}

\ifthenelse{\boolean{arxiv}}{
\bibliographystyle{unsrtnat}
}{
\bibliographystyle{abbrvnat}
}
\bibliography{main}

\clearpage
\onecolumn
\appendix
{\LARGE \textbf{Supplementary}}

\section{Model Implementation Details}
\label{sec:detailimp}
 
Code for our implementation can be found at \url{\repo}.
Here we describe how one can implement our model from scratch
in a deep learning framework.
The main argument in this paper is
that one can apply strong inductive
biases to a deep learning model to simplify the extraction
of a symbolic representation of the learned model.
While we emphasize that
this idea is general, in this section we focus
on the specific Graph Neural Networks we have used as an example throughout
the paper.
 
\subsection{Basic Graph Representation}
 
We would like to use the graph $G=(V, E)$
to predict an updated graph $G'=(V', E)$.
Our input dataset is a graph $G=(V, E)$ consisting of $N^v$ nodes with $L^v$ features each:
$V=\{\mathbf{v}_i\}_{i=1:N^v}$, with each $\mathbf{v}_i \in \mathbb{R}^{L^v}$.
The nodes are connected by $N^e$ edges:
$E=\{(r_k, s_k)\}_{k=1:N^e}$, where $r_k, s_k \in\{1:N^v\}$ are
the indices for the receiving and sending nodes, respectively.
We would like to use this graph to predict
another graph $V'= \{\mathbf{v}_i'\}_{i=1:N^v}$,
where each $\mathbf{v}_i'\in\mathbb{R}^{L^{v\prime}}$ is the node corresponding to $\mathbf{v}_i$.
The number of features in these predicted nodes, $L^{v\prime}$, need not necessarily be the
same as for the input nodes ($L^v$), though this could be the case
for dynamical models where one is predicting updated states of particles.
For more general regression problems, the number of output features is arbitrary.
 
\paragraph{Edge model.}
 
The prediction is done in two parts.
We create the first neural network, the edge model (or ``message function''),
to compute messages from one node to another:
$\phi^e: \mathbb{R}^{L^v}\times \mathbb{R}^{L^v}\rightarrow \mathbb{R}^{L^{e'}}$.
Here, $L^{e'}$ is the number of message features. In the bottleneck model,
one sets $L^{e'}$ equal to the known dimension of the force, which is $2$ or $3$ for us. In our models,
we set $L^{e'}=100$ for the standard and L$_1$ models, and $200$ for the KL model
(which is described separately later on). We create $\phi^e$ as a multi-layer perceptron
with ReLU activations and two hidden layers, each with $300$ hidden nodes.
The mapping is $\mathbf{e}_k' = \phi^e(\mathbf{v}_{r_k}, \mathbf{v}_{s_k})$ for all edges
indexed by $k$ (i.e., we concatenate the receiving and sending node features).
 
\paragraph{Aggregation.} These messages are then pooled via element-wise summation for each receiving node $i$ into
the summed message, $\bar{\mathbf{e}}_i'\in\mathbb{R}^{L^{e'}}$.
This can be written as $\bar{\mathbf{e}}_i' = \sum_{k\in \{1:N^e | r_k = i\}} \mathbf{e}_k'$.
 
\paragraph{Node model.} We create a second neural network
to predict the output nodes, $\mathbf{v}_i'$, for each $i$
from the corresponding summed message and input node.
This net can be written
as $\phi^v:\mathbb{R}^{L^v}\times\mathbb{R}^{L^{e'}}\rightarrow\mathbb{R}^{L^{v\prime}}$,
and has the mapping: $\hat{\mathbf{v}}_i' = \phi^v(\mathbf{v}_{i},  \bar{\mathbf{e}}_i')$,
where $\hat{\mathbf{v}}_i'$ is the prediction for $\mathbf{v}_i'$.
We also create $\phi^v$ as a multi-layer perceptron
with ReLU activations and two hidden layers, each with $300$ hidden nodes.
This model is then trained with the loss function as described later in this section.

\begin{samepage}
\paragraph{Summary.} We can write out our forward model for the bottleneck, standard,
and L$_1$ models as:
\begin{align*}
    \text{Input graph }G&=(V, E)\text{ with}\\
    \text{nodes (e.g., positions of particles) }V&=\{\mathbf{v}_i\}_{i=1:N^v};\ \mathbf{v}_i \in \mathbb{R}^{L^v},\text{ and}\\
    \text{edges (indices of connected nodes) }  E&=\{(r_k, s_k)\}_{k=1:N^e};\ r_k, s_k\in \{1:N^v\}.\\
    \text{Compute messages for each edge: }\mathbf{e}_k' &= \phi^e(\mathbf{v}_{r_k}, \mathbf{v}_{s_k}),\\
    \mathbf{e}_k'\in\mathbb{R}^{L^{e'}},\text{ then}\\
    \text{sum for each receiving node }i:\    \bar{\mathbf{e}}_i' &= \sum_{k\in \{1:N^e | r_k = i\}} \mathbf{e}_k',\\ 
    \bar{\mathbf{e}}_i'\in\mathbb{R}^{L^{e'}}.\\
    \text{Compute output node prediction: }\hat{\mathbf{v}}_i' &= \phi^v(\mathbf{v}_{i},  \bar{\mathbf{e}}_i')\\
    \hat{\mathbf{v}}_i'\in\mathbb{R}^{L^{v\prime}}.
\end{align*}
\end{samepage}

\paragraph{Loss.} We jointly optimize the parameters in $\phi^v$ and $\phi^e$
via mini-batch gradient descent with Adam as the optimizer.
Our total loss function for optimizing is:
\begin{align*}
    \mathcal{L} &= \mathcal{L}_v + \alpha_1 \mathcal{L}_e + \alpha_2\mathcal{L}_n,\text{ where} \\
    \text{the prediction loss is }\mathcal{L}_v &= \frac{1}{N^v}\sum_{i\in\{1:N^v\}} \abs{\mathbf{v}_i' - \hat{\mathbf{v}}_i'},\\
    \text{the message regularization is }\mathcal{L}_e &= \frac{1}{N^e}\left\{\begin{array}{cc}
        \sum_{k\in\{1:N^e\}} \abs{\mathbf{e}_k'},&        \text{L}_1\\
        0,&        \text{Standard}\\
        0,&        \text{Bottleneck}
    \end{array}\right.,\\
    \text{with the regularization constant } \alpha_1&=10^{-2}, \text{ and the}\\
    \text{regularization for the network weights is } \mathcal{L}_n &= \sum_{l=\{1:N^l\}} \abs{w_l}^2,\\
    \text{with }\alpha_2&=10^{-8},
\end{align*}
where  $\mathbf{v}_i'$ is the true value for the predicted node $i$.
$w_l$ is the $l$-th network parameter out of $N^l$ total parameters.
This implementation can be visualized during training in the video \url{\repo}.
During training, we also apply a
random translation augmentation to all
the particle positions to artificially generate more training data.
 
Next, we describe the KL variant of this model.
Note that for the cosmology example in \cref{sec:cosmo},
we use the L$_1$ model described above with $500$ hidden nodes (found with coarse
hyperparameter tuning to optimize accuracy) instead
of $300$, but other parameters are set the same.
 
\subsection{KL Model}
The KL model is a variational version of the GN implementation
above,
which models the messages as distributions. 
We choose a normal distribution for each message component
with a prior of $\mu=0,\ \sigma=1$.
More specifically, the output of $\phi^e$ should now map to twice
as many features as it is predicting a mean and variance,
hence we set $L^{e'}=200$. The first half
of the outputs of $\phi^e$ now represent the means,
and the second half of the outputs represent the log variance of a particular
message component. In other words,
\begin{align*}
    \boldsymbol{\mu}_k' &= \phi^e_{1:100}(\mathbf{v}_{r_k}, \mathbf{v}_{s_k}),\\
    \boldsymbol{\sigma}_k'^2 &= \exp(\phi^e_{101:200}(\mathbf{v}_{r_k}, \mathbf{v}_{s_k})),\\
    \mathbf{e}_{k}' &\sim\mathcal{N}(\boldsymbol{\mu}_k', \text{diag}(\boldsymbol{\sigma}_k'^2)),\\
    \bar{\mathbf{e}}_{i}' &= \sum_{k\in \{1:N^e | r_k = i\}} \mathbf{e}_k',\\
    \hat{\mathbf{v}}_{i}' &= \phi^v(\mathbf{v}_i, \bar{\mathbf{e}}_{i}'),
\end{align*}
where $\mathcal{N}$ is a multinomial Gaussian distribution.
Every time the graph network is run, we calculate the mean and log variance
of messages, sample each message once to calculate $\mathbf{e}_k'$,
and pass those samples through a sum to compute a sample of $\bar{\mathbf{e}}_i'$
and then pass that value through the edge function to compute
a sample of $\hat{\mathbf{v}}_{i}'$. The loss is calculated
normally, except for $\mathcal{L}_e$, which becomes the KL
divergence with respect to our Gaussian prior of $\mu=0,\ \sigma=1$:
$$\mathcal{L}_e = \frac{1}{N^e} \sum_{k=\{1:N^e\}} \sum_{j=\{1:L^{e'}/2\}}\frac{1}{2}\left(\mu^{\prime 2}_{k, j} + \sigma^{\prime 2}_{k,j} - \log(\sigma_{k,j}^{\prime 2})\right),$$
with $\alpha_1=1$ (equivalent to $\beta=1$ for the loss of a $\beta$-Variational
Autoencoder; simply the standard VAE). The KL-divergence loss also encourages sparsity in the messages
$\mathbf{e}_k'$ similar to the L$_1$ loss. The difference
is that here, an uninformative message component will have $\mu=0, \sigma=1$ (a KL of 0)
rather than a small absolute value.
We train the networks with a decaying learning schedule as given in the example code.

\subsection{Constraining Information in the Messages}
The hypothesis which motivated our graph network inductive bias
is that if one minimizes the dimension of the vector space used
by messages in a GN, the components of message vectors
will learn to be linear combinations of the true forces (or equivalent underlying summed function)
for the system being learned.
The key observation is that $\mathbf{e}'_k$ could learn to correspond to the true force vector imposed on the $r_k$-th body due to its interaction with the $s_k$-th body.
 
Here, we sketch a rough mathematical explanation of our hypothesis that we will
reconstruct the true force in the graph network given our inductive biases.
Newtonian mechanics prescribes that force vectors, $\mathbf{f}_k \in \mathcal{F}$, can be summed to produce a net force, $\sum_k \mathbf{f}_k = \mathbf{\bar{f}} \in \mathcal{F}$, which can then be used to update the dynamics of a body. 
Our model uses the $i$-th body's pooled messages, $\mathbf{\bar{e}}'_i$ to update the body's state via $\mathbf{v}_i' = \phi^v(\mathbf{v}_i, \mathbf{\bar{e}}'_i)$.
 If we assume our GN is trained to predict accelerations perfectly for any number of bodies, this means (ignoring mass) that $\mathbf{\bar{f}}_i = \sum_{r_k = i} \mathbf{f}_k = \phi^v(\mathbf{v}_i, \sum_{r_k = i}  \mathbf{e}'_k) = \phi^v(\mathbf{v}_i, \mathbf{\bar{e}}'_i)$.
Since this is true for any number of bodies, we also have the result for a single interaction: $\mathbf{\bar{f}}_i = \mathbf{f}_{k, r_k=i} = \phi^v(\mathbf{v}_i,\mathbf{e}'_{k, r_k=i}) = \phi^v(\mathbf{v}_i,\mathbf{\bar{e}}'_i)$. Thus, we can substitute this expression into the multi-interaction case: $\sum_{r_k=i}\phi^v(\mathbf{v}_i,\mathbf{e}'_k) = \phi^v(\mathbf{v}_i,\mathbf{\bar{e}}'_i) = \phi^v(\mathbf{v}_i,\sum_{r_k=i}\mathbf{e}'_k)$.
From this relation, we see that $\phi^v$ has to be a linear transformation conditioned on $\mathbf{v}_i$.
Therefore, for cases where $\phi^v(\mathbf{v}_i, \mathbf{\bar{e}}'_i)$ is invertible in $\mathbf{\bar{e}}'_i$
(which becomes true when $\mathbf{\bar{e}}'_i$ is the same dimension as the output of $\phi^v$),
we can write $\mathbf{e}'_k = (\phi^v(\mathbf{v}_i, \cdot))^{-1}(\mathbf{f}_k)$,
which is also a linear transform, meaning that
the message vectors are linear transformations of the true forces when $L^{e'}$ is equal to the dimension of the forces.
 
If the dimension of the force vectors (or what the minimum dimension
of the message vectors ``should'' be)
is unknown, one can encourage the messages to be sparse
by applying L$_1$ or Kullback-Leibler regularizations to the messages
in the GN.
The aim is for the messages to learn the minimal vector space
required for the computation automatically. 
This is a more mathematical explanation of why the message features
are linear combinations of the force vectors, when our inductive
bias of a bottleneck or sparse regularization is applied.
We emphasize that this is a new contribution: never before has previous work explicitly identified the forces in a graph network.
 
\paragraph{General Graph Neural Networks.} In all of our models here,
we assume the dataset does not have edge-specific features, such as a different
coupling constants between different particles, but these
could be added by concatenating edge features to the receiving
and sending node input to $\phi^e$.
We also assume there are no global properties. The graph neural network
is described in general form in \cite{battaglia2018relational}.
All of our techniques are applicable to the general form: one would
approximate $\phi^e$ with a symbolic model with included input edge parameters,
and also fit the global model, denoted $\phi^u$.

\subsection{Flattened Hamiltonian Graph Network.}
\label{sec:appham}
 
As part of this study,
we also consider an alternate dynamical model that
is described by a linear latent space
other than force vectors.
In the Hamiltonian formalism of classical mechanics, energies
of pairwise interactions and kinetic and potential energies of particles
are pooled into a global energy value, $\H$, which is a scalar.
We label pairwise interaction energy $\Hp$
and the energy of individual particles as $\Hs$.
Thus, using our previous graph notation,
we can write the total energy of a system as:
\begin{equation}\label{eqn:totalenergy}
    \H = \sum_{i=1:N^v}\Hs(\v{v}_i) + \sum_{k\in\{1:N^e\}} \Hp(\v{v}_{r_k}, \v{v}_{s_k}).
\end{equation}
For particles interacting via gravity, this would be
\begin{equation}
\H = \sum_{i}\frac{p_i^2}{2m_i} - \frac{1}{2}\sum_{i\neq j}\frac{m_i m_j}{\abs{\v{r}_i - \v{r}_j}},
\end{equation}
where $\v{p}_i, m_i, \v{r}_i$ indicates the momentum, mass, and position
of particle $i$, respectively, and we have set the gravitational constant to $1$.
Following \citet{greydanus,sanchez}, we could model $\H$ as a neural network,
and apply Hamilton's equations to create a dynamical model.
More specifically, as in \citet{sanchez},
we can predict $\H$ as the global property of a GN (this is called a Hamiltonian
Graph Network or HGN).
However, energy, like forces in Cartesian
coordinates, is a summed quantity. In other words,
energy is another ``linear latent space''
that describes the dynamics.
 
Therefore, we argue that an HGN will be more interpretable
if we explicitly sum up energies over the system, rather than compute $\H$
as a global property of a GN.
Here, we introduce the ``Flattened Hamiltonian Graph Network,'' or ``FlatHGN'',
which uses \cref{eqn:totalenergy} to construct
a model that works on a graph.
We set up two Multi-Layer Perceptrons (MLPs), one for each node:
\begin{equation}
    \Hs:\mathbb{R}^{L^v} \rightarrow \mathbb{R},
\end{equation}
and one for each edge:
\begin{equation}
    \Hp:\mathbb{R}^{L^v}\times\mathbb{R}^{L^v} \rightarrow \mathbb{R}.
\end{equation}
Note that the derivatives of $\H$ now propagate through the pool, e.g.,
\newcommand\tmpH[1]{\frac{\partial #1}{\partial \v{v}_i}}
\begin{align}
    \tmpH{\H(V)} &= \tmpH{\Hs(\v{v}_i)} + \sum_{r_k=i} \tmpH{\Hp(\v{e}_k, \v{v}_{r_k}, \v{v}_{s_k})}\\
    &\hphantom{xxx} + \sum_{s_k=i} \tmpH{\Hp(\v{e}_k, \v{v}_{r_k}, \v{v}_{s_k})}. \nonumber
\end{align}
This model is similar to the Lagrangian Graph Network
proposed in \citet{lagrangian}.
Now, should this FlatHGN learn energy functions such that we can successfully
model the dynamics of the system with Hamilton's equations, we would expect
that $\Hs$ and $\Hp$ should be
analytically similar to parts of the true Hamiltonian.
Since we have broken the traditional HGN into a FlatHGN, we now
have pairwise and self energies, rather than a single global
energy, and these are simpler to extract and interpret.
This is a similar inductive bias to the GN we introduced
previously. To train a FlatHGN, one can follow our strategy above,
with the output predictions made using
Hamilton's equations applied to our $\H$.
One difference is that we also regularize $\Hp$, 
since it is degenerate with $\Hs$ in that it can
pick up self energy terms.

\section{Simulations}
 
\label{sec:simulations}
\begin{figure*}[h]
    \centering
    \begin{subfigure}[b]{0.33\textwidth}
    \centering
    \includegraphics[width=\textwidth]{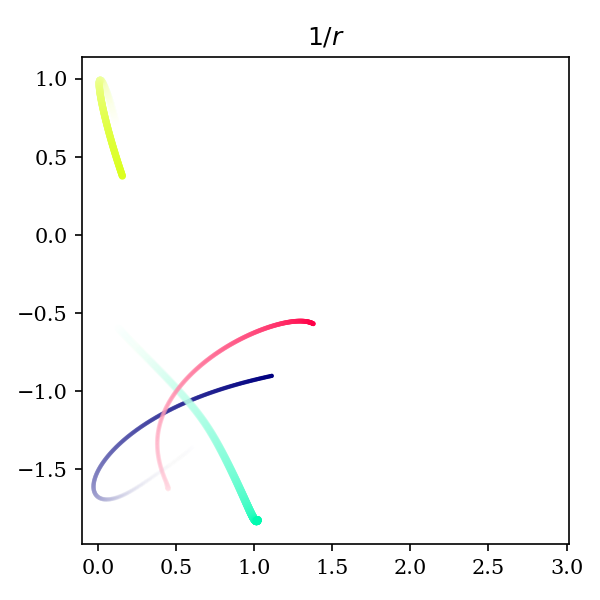}
    \end{subfigure}%
    \begin{subfigure}[b]{0.33\textwidth}
    \centering
    \includegraphics[width=\textwidth]{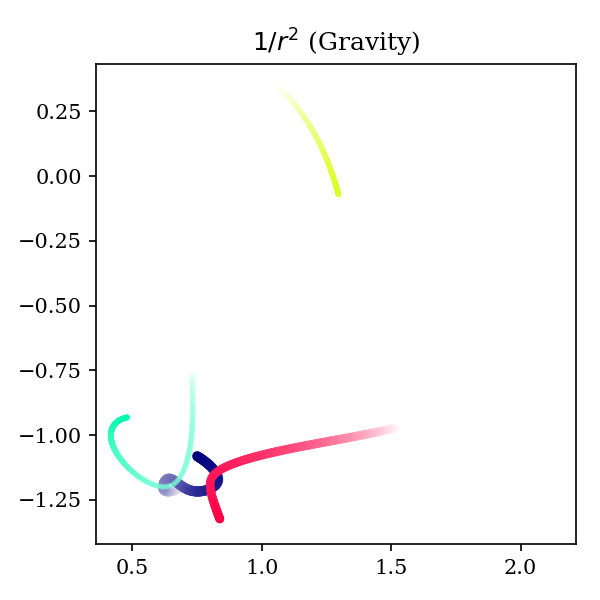}
    \end{subfigure}%
    \begin{subfigure}[b]{0.33\textwidth}
    \centering
    \includegraphics[width=\textwidth]{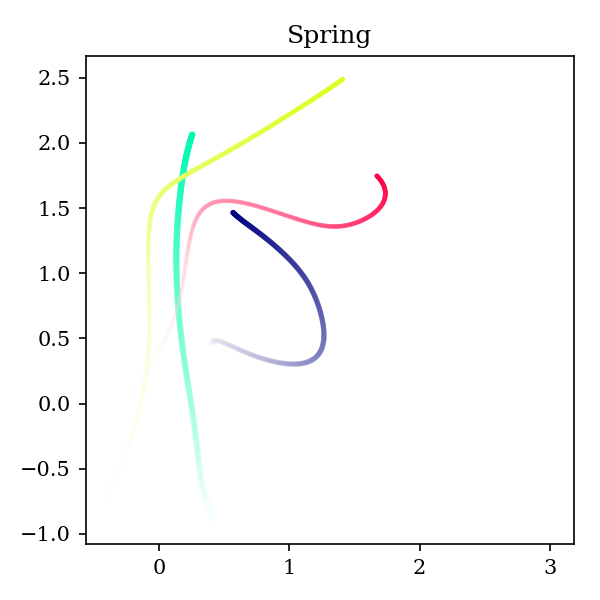}
    \end{subfigure}
    \begin{subfigure}[b]{0.33\textwidth}
    \centering
    \includegraphics[width=\textwidth]{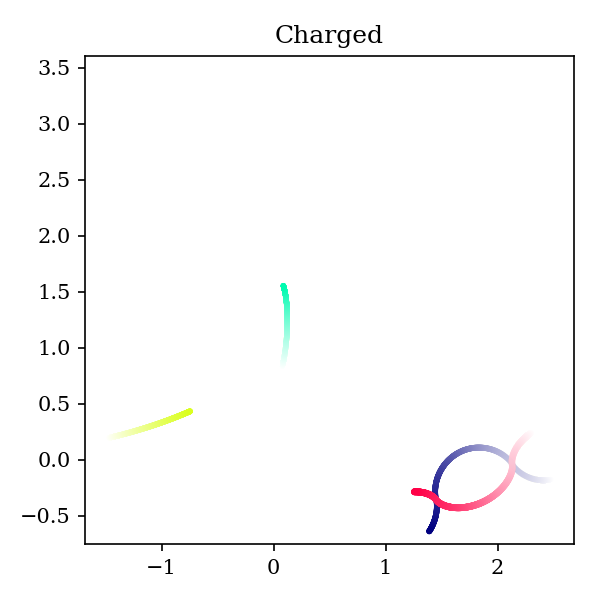}
    \end{subfigure}%
    \begin{subfigure}[b]{0.33\textwidth}
    \centering
    \includegraphics[width=\textwidth]{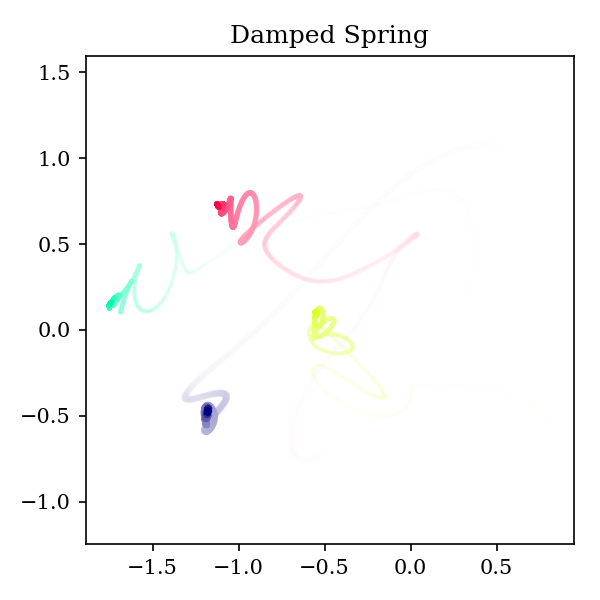}
    \end{subfigure}%
    \begin{subfigure}[b]{0.33\textwidth}
    \centering
    \includegraphics[width=\textwidth]{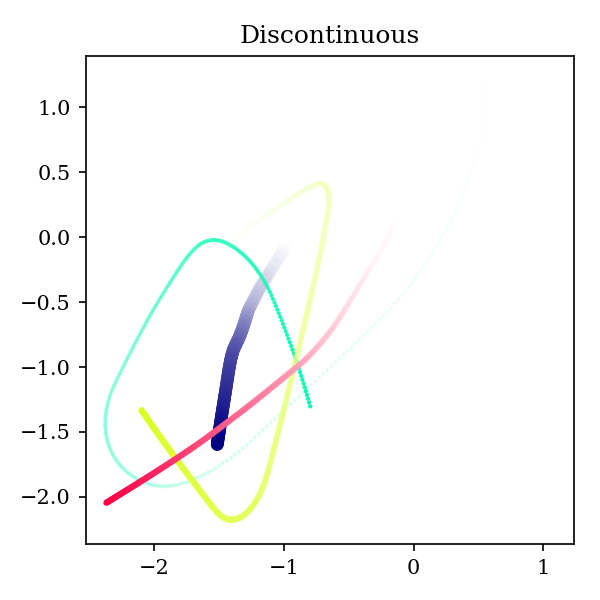}
    \end{subfigure}%
    \caption{Examples of a selection of simulations,
    for 4 nodes and two dimensions. Decreasing transparency
    shows increasing time, and size of points shows mass.
    }
   \label{fig:simulations}
\end{figure*}

Our simulations for \cref{sec:force,sec:hamiltonian} were written using the JAX
library (\url{https://github.com/google/jax}) so that
we could easily vectorize computations over the entire
dataset of 10,000 simulations.
Example ``long exposures'' for each simulation
in 2D are shown in \cref{fig:simulations}.
To create each simulation, we set up the following potentials
between two particles, $1$ (receiving) and $2$ (sending).
Here, $r'_{12}$ is the distance
between two particles plus $0.01$ to prevent singularities.
For particle $i$,
$m_i$ is the mass, $q_i$ is the charge,
$n$ is the number of particles in the simulation,
$\v{r}_i$ is the position of a particle, and
$\dot{\v{r}}_i$ is the velocity of a particle.
\begin{align*}
1/r^2:~ U_{12} &= 
- m_1 m_2 / r'_{12} \\
1/r:~ U_{12} &= 
m_1 m_2 \log(r'_{12}) \\
\text{Spring}:~ U_{12} &= 
(r'_{12} - 1)^2 \\
\text{Damped}:~ U_{12} &= 
    (r'_{12} - 1)^2 + \v{r}_1\cdot \dot{\v{r}}_1/n\\
\text{Charge}:~ U_{12} &= 
    q_1 q_2 / r'_{12}\\
\text{Dicontinuous}:~ U_{12} &= 
    \left\{
        \begin{array}{lr}
            0, &  r'_{12} < 2\\
            (r'_{12}-1)^2, &r'_{12} \geq 2
        \end{array}
        \right.
\end{align*}
All variables lack units.
Here, $m_i$ is sampled from
a log-normal distribution
with $\mu=0, \sigma=1$.
Each component of
$\v{r}_i$
and $\dot{\v{r}}_i$
is randomly sampled
from a normal distribution
with $\mu=0, \sigma=1$.
$q_i$ is randomly
drawn from a set of two elements: $\{-1, 1\}$, representing
charge.
The acceleration of a given particle is then
\begin{equation}
    \ddot{\v{r}}_i = -\frac{1}{m_i} \sum_j \nabla_{\v{r}_i} U_{ij}.
\end{equation}
This is integrated over 1000 time steps of a fixed
step size for a given random initial configuration
using an adaptive RK4 integrator.
The step size varies for each simulation due
to the differences in scale. It is:
0.005 for $1/r$,
0.001 for $1/r^2$,
0.01 for Spring,
0.02 for Damped,
0.001 for Charge,
and 0.01 for Discontinuous.
Each simulation is performed
in two and three dimensions,
for 4 and 8 bodies. We store these
simulations on disk. For training,
the simulations for the particular problem being
studied are loaded, and
each instantaneous snapshot
of each simulation is converted to a fully connected graph,
with the predicted property (nodes of $V'$, see \cref{sec:detailimp}) being the acceleration
of the particles at that snapshot.
 
The test loss of each model trained on each simulation set
is given in \cref{tbl:losses}.
\begin{table*}[h]
\centering
\begin{tabular}{@{}lccccc@{}}
\toprule
Sim. & Standard & Bottleneck & L$_1$       &  KL      & FlatHGN \\ \midrule
Charge-2 & \textbf{49} & 50 & 52 & 60 & 55 \\
Charge-3 & 1.2 & 0.99 & \textbf{0.94} & 4.2 & 3.5 \\
Damped-2 & \textbf{0.30} & 0.33 & \textbf{0.30} & 1.5 & 0.35 \\
Damped-3 & 0.41 & 0.45 & \textbf{0.40} & 3.3 & 0.47 \\
Disc.-2 & 0.064 & 0.074 & \textbf{0.044} & 1.8 & 0.075 \\
Disc.-3 & 0.20 & 0.18 & \textbf{0.13} & 4.2 & 0.14 \\
$r^{-1}$-2 & 0.077 & 0.069 & 0.079 & 3.5 & \textbf{0.05} \\
$r^{-1}$-3 & 0.051 & 0.050 & 0.055 & 3.5 & \textbf{0.017} \\
$r^{-2}$-2 & 1.6 & 1.6 & \textbf{1.2} & 9.3 & 1.3 \\
$r^{-2}$-3 & 4.0 & 3.6 & 3.4 & 9.8 & \textbf{2.5} \\
Spring-2 & 0.047 & 0.046 & 0.045 & 1.7 & \textbf{0.016} \\
Spring-3 & 0.11 & 0.11 & 0.090 & 3.8 & \textbf{0.010} \\ \bottomrule
\end{tabular}
\caption{Test prediction losses for each model on each dataset in two and three
dimensions.
The training was done with the same batch size, schedule,
and number of epochs. }
\label{tbl:losses}
\end{table*}
 
\begin{figure*}[h]
    \centering
    \begin{subfigure}[b]{0.45\textwidth}
    \centering
        \includegraphics[width=\textwidth]{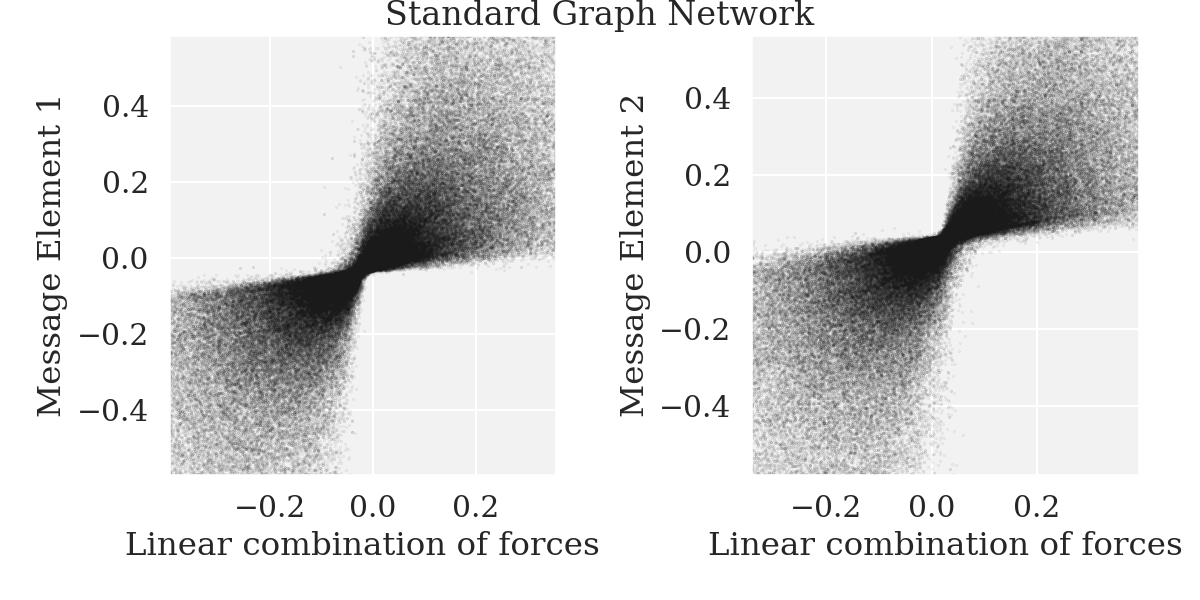}
    \end{subfigure}%
    \begin{subfigure}[b]{0.45\textwidth}
    \centering
    \includegraphics[width=\textwidth]{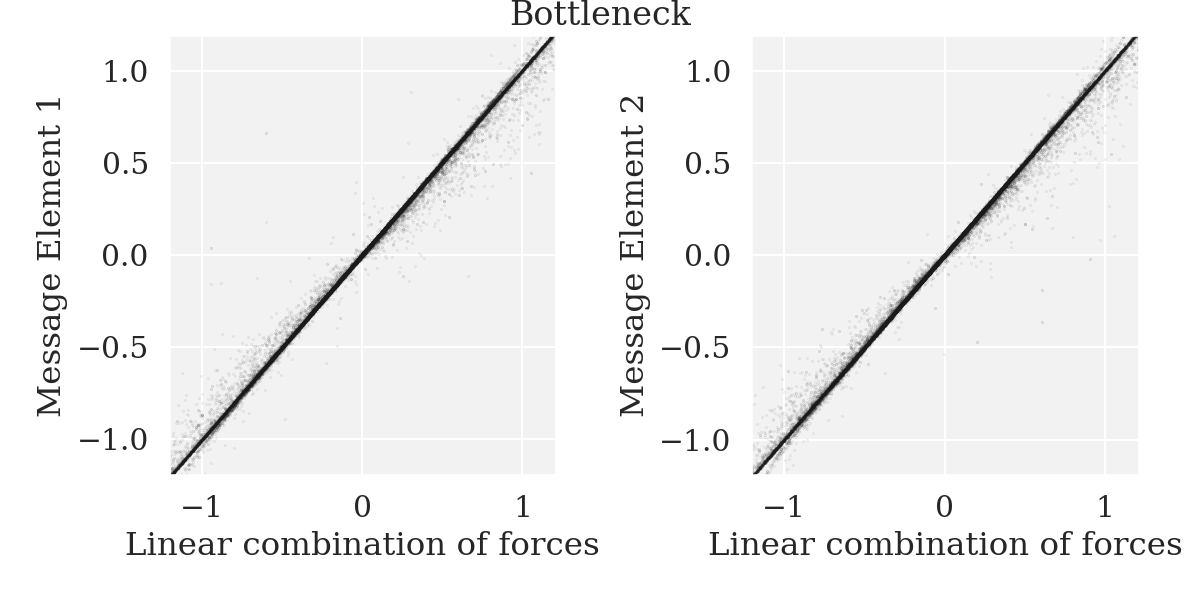}
    \end{subfigure}
    \begin{subfigure}[b]{0.45\textwidth}
    \centering
        \includegraphics[width=\textwidth]{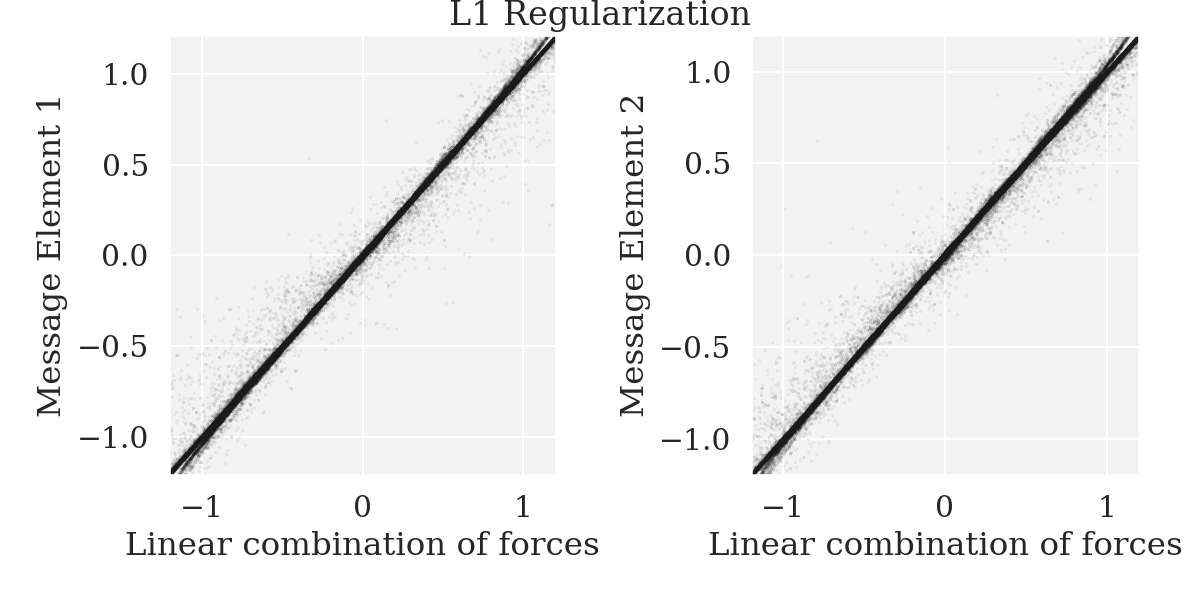}
    \end{subfigure}%
    \begin{subfigure}[b]{0.45\textwidth}
    \centering
    \includegraphics[width=\textwidth]{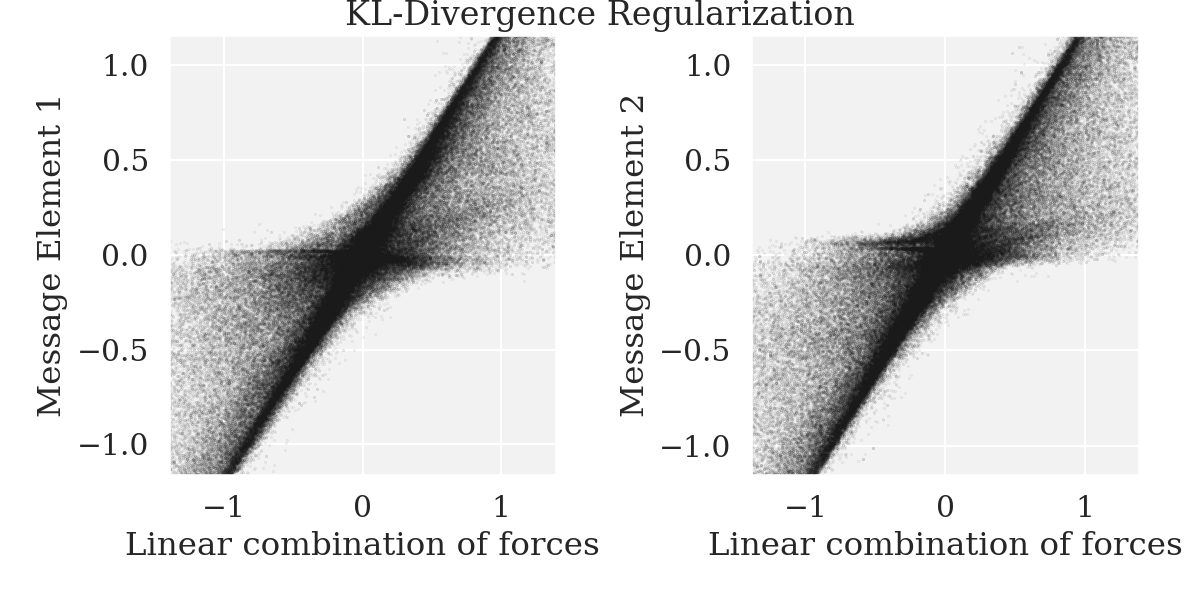}
    \end{subfigure}
    \caption{
        The most significant message components of each model compared
        with a linear combination of the force components: this example,
  the spring simulation in 2D with eight nodes for training.
        These plots demonstrate that the GN's messages
        have learned to be linear transformations
        of the vector components of the true force, in this case a springlike
        force, after applying an inductive bias to the messages.}
   \label{fig:fitted}
\end{figure*}

As described in the text (and visualized in the drive video), we can fit 
linear combinations of the true force components to each of the significant
features of a message vector. This fit is summarized by \cref{tbl:forcefit},
and the fit itself is visualized in \cref{fig:fitted} for various
models on the 2D spring simulation.

\section{Symbolic Regression Details}
After training a model on each simulation,
we convert a deep learning model to a symbolic expression by approximating
subcomponents of the model with symbolic regression,
over observed inputs and outputs.
For our aforementioned GNN implementation, we can record
the outputs of $\phi^e$ and $\phi^v$ for various data points
in the training set.
 
For models other than the bottleneck and Hamiltonian model (where we explicitly limit the features) we calculate
the most significant output features of $\phi^e$ (we also refer
to the output features as ``message components''). For the
L$_1$ and standard model, this is done by sorting the message components
with the largest standard deviation; the most significant feature
is the one with the largest standard deviation, which are the features
we study. For the KL model, we consider the feature
with the largest KL-divergence: $\mu^2 + \sigma^2 - \log(\sigma^2)$.
These features are the ones we consider to be containing information
used by the GN, so are the ones we fit symbolic expressions to.
 
As an example, here we fit the most significant feature,
which we refer to as $\phi^e_1$, over random examples of the training dataset.
We do this for the particle simulations in \cref{sec:force}.
The inputs to the actual $\phi^e_1$ neural network are:
$m_1, m_2, q_1, q_2, x_1, x_2, \ldots$ (mass, charge, and Cartesian positions
of receiving and sending node), leaving us with many examples of
$(m_1, m_2, q_1, q_2, x_1, x_2, \ldots,  \phi^e_1)$. We would like to
fit a symbolic expression to map
$(m_1, m_2, q_1, q_2, x_1, x_2, \ldots)\rightarrow \phi^e_1$.
To simplify things for this symbolic model, we convert the input position variables
to a more interpretable format: $\Delta x=x_2-x_1$ for $x$ displacement, likewise
for $y$ (and $z$, if it is a 3D simulation),
and $r = \sqrt{\Delta x^2 + \Delta y^2 \ (+\Delta z^2)}$ for distance.
 
\begin{figure*}[h]
    \centering
    \includegraphics[width=0.60\textwidth]{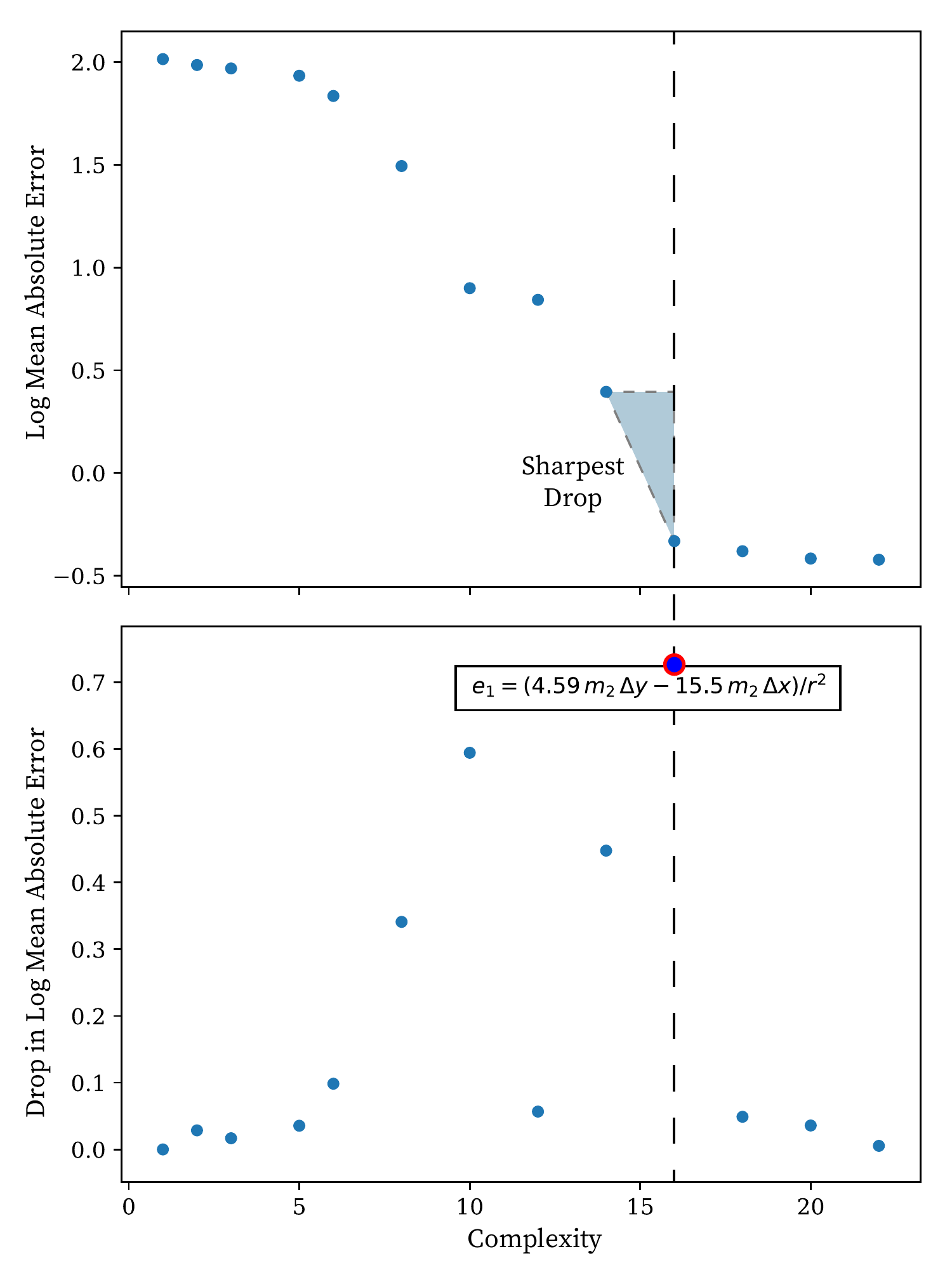}
    \caption{A plot 
    of the data for the $1/r$ simulation in \cref{tbl:square},
    indicating mean absolute error versus complexity
    in the top plot and fractional drop in mean absolute error over the next-best
    model in the bottom plot.
    As indicated, we take the largest drop in log-loss
    over a single increase in complexity
    as the chosen model---it is our
    parametrization of Occam's Razor.}
    \label{fig:r_complexity}
\end{figure*}

We then pass these $(m_1, m_2, q_1, q_2, \Delta x, \Delta y, (\Delta z,) r, \phi^e_1)$ examples (we take
5000 examples for each of our tests)
to \eureqa, and ask it to fit $\phi^e_1$ as a function of the others by minimizing the mean absolute error (MAE).
We allow it to use the operators $+, -, \times, /, >, <,$ \textasciicircum$, \exp,
\log, \textsc{IF}(\cdot,\cdot,\cdot)$ as well as
real constants in its solutions.
We score complexity 
by counting the number of occurrences of each operator, constant,
and input variable.
We weight \textasciicircum$, \exp,
\log, \textsc{IF}(\cdot,\cdot,\cdot)$
as three times the other operators,
since these are more complex operations. 
\eureqa outputs the best equation at each 
complexity level, denoted by $c$. Example outputs
are shown in \cref{tbl:square} for the $1/r$ and $1/r^2$
simulations. We select a formula from this list by taking
the one that maximizes the fractional drop
in mean absolute error (MAE) over 
an increase in complexity from the next best model.
This is analogous to Occam's Razor: we jointly
optimize for simplicity and accuracy of the model.
The objective itself can be written as maximizing
$(-\Delta \log(\text{MAE}_c)/\Delta c)$ over the best
model at each maximum complexity level,
and is schematically illustrated in \cref{fig:r_complexity}.
We find experimentally that this score
produces the best-recovered solutions in a variety
of tests on different generating equations.

Following the process of fitting analytic
equations to the messages, we fit a single analytic expression
to model $\phi^e_1$ as a function of the simplified input variables.
We recover many analytical expressions that were used to generate the data,
examples of which are listed below ($a, b$ indicate learned constants):
\begin{itemize}
    \item Spring, 2D, L$_1$ (expect $\phi^e_1\approx (\v{a}\cdot (\Delta x, \Delta y)) (r - 1) + b$). 
$$\phi^e_1\approx 1.36 \Delta y + 0.60 \Delta x - \frac{0.60 \Delta x + 1.37 \Delta y}{r} - 0.0025 $$
    \item $1/r^2$, 3D, Bottleneck (expect $\phi^e_1\approx \frac{\v{a}\cdot (\Delta x, \Delta y, \Delta z)}{r^3} + b$). 
$$\phi^e_1\approx \frac{0.021\Delta x m_2 - 0.077 \Delta y m_2}{r^3}$$
    \item Discontinuous, 2D, L$_1$ (expect $\phi^e_1\approx \textsc{IF}(r>2, (\v{a}\cdot (\Delta x, \Delta y, \Delta z)) r, 0) + b$). 
            $$\phi^e_1 \approx \textsc{IF}(r>2,
            0.15 r \Delta y + 0.19 r \Delta x,
            0) - 0.038$$
\end{itemize}

\begin{table}[h]
\centering
\begin{tabular}{@{}lll@{}}
\toprule
Solutions extracted for the 2D $1/r^2$ Simulation                                                                    & MAE & Complexity \\ \midrule
$\phi^e_1 = 0.162 + (5.62 + 20.3m_2\Delta x - 153m_2\Delta y)/r^3$ & 17.954713          & 22         \\
$\phi^e_1 = (6.07 + 19.9m_2\Delta x - 154m_2\Delta y)/r^3$         & 18.400224          & 20         \\
$\phi^e_1 = (3.61 + 20.9\Delta x - 154m_2\Delta y)/r^3$            & 42.323236          & 18         \\
$\phi^e_1 = (31.6\Delta x - 152m_2\Delta y)/r^3$                   & 69.447467          & 16         \\
$\phi^e_1 = (2.78 - 152m_2\Delta y)/r^3$                           & 131.42547          & 14         \\
\rowcolor{Gray} $\phi^e_1 = -142m_2\Delta y/r^3$                                   & 160.31243          & 12  \\
$\phi^e_1 = -184\Delta y/r^2$                                      & 913.83751          & 8          \\
$\phi^e_1 = -7.32\Delta y/r$                                                        & 1520.9493          & 6          \\
$\phi^e_1 = -0.282m_2\Delta y$                                                      & 1551.3437          & 5          \\
$\phi^e_1 = -0.474\Delta y$                                                         & 1558.9756          & 3          \\
$\phi^e_1 = 0.0148$                                                                 & 1570.0905          & 1         \\ \bottomrule
\toprule
Solutions extracted for the 2D $1/r$ Simulation & MAE & Complexity \\ \midrule
$\phi^e_1 = (4.53m_2\Delta y - 1.53\Delta x - 15.0m_2\Delta x)/r^2 - 0.209$ & 0.37839388         & 22         \\ 
$\phi^e_1 = (4.58m_2\Delta y - \Delta x - 15.2m_2\Delta x)/r^2 - 0.227$  & 0.38 & 20         \\ %
$\phi^e_1 = (4.55m_2\Delta y - 15.5m_2\Delta x)/r^2 - 0.238$ & 0.42 & 18         \\
\rowcolor{Gray} $\phi^e_1 = (4.59m_2\Delta y - 15.5m_2\Delta x)/r^2$ & 0.46575519         & 16   \\
$\phi^e_1 = (10.7\Delta y - 15.5m_2\Delta x)/r^2$                           & 2.48 & 14         \\
$\phi^e_1 = (\Delta y - 15.6m_2\Delta x)/r^2$                                & 6.96 & 12         \\
$\phi^e_1 = -15.6m_2\Delta x/r^2$                                           & 7.93 & 10         \\
$\phi^e_1 = -34.8\Delta x/r^2$                                              & 31.17 & 8          \\
$\phi^e_1 = -8.71\Delta x/r$                                                                 & 68.345174          & 6          \\
$\phi^e_1 = -0.360m_2\Delta x$                                                               & 85.743106          & 5          \\
$\phi^e_1 = -0.632\Delta x$                                                                  & 93.052677          & 3          \\
$\phi^e_1 = -\Delta x$                                                                        & 96.708906          & 2          \\
$\phi^e_1 = -0.303$                                                                          & 103.29053          & 1          \\
    \bottomrule
\end{tabular}
\caption{Results of using symbolic regression to fit
equations to the most significant (see text) feature of
$\phi^e$, denoted $\phi^e_1$, for the $1/r^2$ (top) and $1/r$ (bottom) force laws,
extracted from the bottleneck model.
We expect to see
$\phi^e_1\approx \frac{\v{a}\cdot (\Delta x, \Delta y, \Delta z)}{r^\alpha} + b$,
    for arbitrary $\v{a}$ and $b$, and $\alpha=2$ for the $1/r$ simulation and
    $\alpha=3$ for the $1/r^2$ simulation, which is approximately what we recover.
    The row with a gray background has the largest fractional drop in mean absolute error
    in their tables, which according
    to our parametrization of Occam's razor, represents the best model.
This demonstrates a technique for learning
an unknown ``force law'' with a constrained graph neural network.}
\label{tbl:square}
\end{table}
 
{}

\paragraph{Examples of failed reconstructions.}
Note that reconstruction does not always succeed, especially
for training strategies other than L$_1$ or bottleneck models that cannot successfully find compact representations of the right dimensionality. We demonstrate some failed
examples below:
\begin{itemize}
    \item Spring, 3D, KL (expect $\phi^e_1\approx (\v{a}\cdot (\Delta x, \Delta y, \Delta z)) (r - 1) + b$). 
        $$\phi^e_1\approx 0.57 \Delta y + 0.32 \Delta z$$
    \item $1/r$, 3D, Standard (expect $\phi^e_1\approx \frac{\v{a}\cdot (\Delta x, \Delta y, \Delta z)}{r^2} + b$). 
            $$\phi^e_1 \approx \frac{0.041 + m_2 \textsc{IF}(\Delta z>0, 0.021, 0.067)}{r} 
            $$
\end{itemize}
We do not attempt to make any general statements about when
symbolic regression applied to the message components
will fail or succeed in extracting the true law.
Simply, we show that it is possible,
for a variety of physical systems, and argue that reconstruction
is more likely by the inclusion of a strong inductive bias in the network.

\begin{table}[h]
    \centering
\begin{tabular}{@{}lcccc@{}}
    \toprule
    Sim. &         Standard &                Bottleneck &                   L$_1$       &                   KL
    \\ \midrule
    Charge-2 &           \xmark &                \cmark  &                   \xmark       &                  \xmark
    \\
    Charge-3 &               \xmark &                \cmark   &                   \xmark       &                   \xmark
    \\
    $r^{-1}$-2 &        \xmark  &                \cmark  &                   \cmark       &                   \cmark
    \\
    $r^{-1}$-3 &          \xmark  &                \cmark  &                   \cmark       &                   \cmark
    \\
    $r^{-2}$-2 &          \xmark &                \cmark  &                   \cmark       &                   \xmark
    \\
    $r^{-2}$-3 &          \xmark &                \cmark  &                   \cmark       &                   \xmark
    \\
    Spring-2 &         \xmark &                \cmark &                   \cmark       &                   \cmark
    \\
    Spring-3 &        \xmark &                \cmark &                   \cmark       &                   \cmark\\ \bottomrule
    \\
\end{tabular}
\caption{Success/failure of a reconstruction of the force law by symbolic regression,
corresponding to the values in \cref{tbl:forcefit}.}
\label{tbl:recons}
\end{table}
A full table of successes and failures in reconstructing the force
law over the different n-body experiments is given in
\cref{tbl:recons}.
While the equations given throughout the paper
were generated with \eureqa,
to create this table in particular, we
switched from \eureqa to \pysr.
This is because \pysr allows us to configure a controlled
experiment with fixed hyperparameters and total mutation steps
for each force law,
whereas Eureqa makes these controls inaccessible.
However, given enough training time, we found \eureqa
and \pysr produced equivalent results
for equations at this simplicity level.

\paragraph{Pure \eureqa experiment}

To demonstrate that \eureqa by itself is not capable
of finding many of the equations considered from the raw
high-dimensional dataset,
we ran it on the simulation data without our GN's factorization
of the problem, giving it the features of every particle.
As expected, even after convergence, it cannot find meaningful equations;
all of the solutions it provides for the n-body system are very poor fits.
One such example of an equation,
for the acceleration of particle $2$ along the $x$ direction
in a $6$-body system under a $1/r^2$ force law, is:
\begin{align*}
    \ddot x_{2} &= \frac{0.426}{367 y_4 - 1470} +
\frac{2.88\times10^5 x_1}{2.08\times10^3 y_4 + 446 y_4^2} 
- 5.98\times10^{-5} x_6 - 109 x_1,
\end{align*}
where the indices refer to particle number.
Despite \eureqa converging,
this equation is evidently meaningless and achieves
a poor fit to the data.
Thus, we argue that raw symbolic regression
is intractable for the problems we consider, and only
after factorization with a neural network do these
problems become feasible
for symbolic regression.

\paragraph{Discovering potentials using FlatHGN.}
Lastly, we also show an example of a successful reconstruction of a
pairwise Hamiltonian from data. We treat the $\Hp$ just as we would
$\phi^e_1$, and fit it to data. The one difference here is that there
are potential $\Hp$ values offset by a constant function of the non-dynamical
parameters (fixed properties like mass) which still produce the correct dynamics,
since only the derivatives of $\Hp$ are used. Thus, we cannot simply
fit a linear transformation of the true $\Hp$ to data to verify it
has learned our generating equation: we must rely
on symbolic regression to extract the full functional form.
We follow the same procedure as before,
and successfully extract the potential for a charge simulation:
$$\Hp\approx \frac{0.0019 q_1 q_2}{r} - 0.0112 - 0.00143 q_1 - 0.00112 q_1 q_2,$$
where we expect $\Hp \approx a \frac{q_1 q_2}{r} + f(q_1, q_2, m_1, m_2)$,
for constant $a$ and arbitrary function $f$,
which shows that the neural network has learned the correct form
of the Hamiltonian.
 
\paragraph{Hyperparameters.}
Since the hyperparameters used internally by
\eureqa are opaque and not tunable, here we discuss the parameters
used in \pysr \citet{pysr},
which are common among many symbolic regression tools.
At a given step of the training, there is a set of active
equations in the ``population''. The number of active equations
is a tunable hyperparameter, and is related to the diversity of the
discovered equations, as well as the number of compute cores being used.
The max size of equations controls the maximum complexity considered,
and can be controlled to prevent the algorithm from wasting cycles on over-complicated
equations.
The operators used in the equations depends on the specific
problem considered, and is another hyperparameter specified by the user.
Next, there is a set of tunable
probabilities associated with each mutation:
how frequently to mutate
an operator into a different operator,
add an operator with arguments, replace an operator and its arguments
with a constant, and so on. 
In some approaches such as with \pysr, the best equations found over the course
of training are randomly reintroduced back into the population. The frequency
at which this occurs is controlled by another hyperparameter.

\section{Video Demonstration and Code}
We include a video demonstration of the central ideas
of our paper at \url{\repo}.
It shows the message components
of a graph network converging to be equal to a linear combination
of the force components when L$_1$ regularization is applied.
Time in each clip of the video is correlated with training epoch.
In this video, the top left corner of the fully revealed
plot corresponds to a single 
test simulation that is 300 time steps long. Four
particles of different masses are initiated with random positions
and velocities, and evolved according to the potential
of a spring with an equilibrium position
of $1$: $(r-1)^2$, where $r$ is the distance
between two particles. The evaluation trajectories
are shown on the right, with the gray particles
indicating the true locations.
The 15 largest message components in terms of standard
deviation over a test set are represented in a sorted
list below the graph network in gray, where darker
color corresponds to a larger standard deviation. Since
we apply L$_1$ regularization to the messages, we expect
this list to grow sparser over time, which it does.
Of these messages, the two largest components are extracted,
and each is fit to a separate
linear combination of the true force components (bottom left).
A better fit to the true force components --- indicating that the messages
represent the force --- are indicated by dots (each dot is a single message)
that lie closer along the $y=x$ line in the bottom middle two scatter
plots.
 
As can be seen in the video, as the messages grow increasingly sparse,
the messages eventually converge to be almost exactly linear combinations
of the true forces. Finally, once the loss is converged,
we also fit symbolic regression to the largest message component.
The video was created using the same training procedure as used
in the rest of the paper. The dataset that the L$_1$ model 
was trained on is the 4-node Spring-2.
Finally, we include the full code required to generate
the animated clips in the above figure. This code contains all of the
models and simulators used in the paper,
along with the default training
parameters. This code can also be accessed in the drive.

\section{Cosmological Experiments}
 
For the cosmological data graph network,
we do a coarse hyperparameter tuning based on predictions of $\delta_i$
and select a GN with 500 hidden units, two hidden layers
per node function and message function. We choose 100 message
dimensions as before.
We keep other hyperparameters
the same as before: L$_1$ regularization with a regularization scale
of $10^{-2}$.
 
Remarkably, the vector space discovered by this graph network is 1 dimensional. 
This is indicated by the fact that only one message component has standard deviation of about $10^{-2}$
and all other 99 components have a standard deviation of under $10^{-8}$.
This suggests that the $\delta_i$ prediction is a sum over some
function of the center halo and each neighboring halo. Thus, we can rewrite
our model as a sum over a function $\phi^e_1$ which takes the
central halo and each neighboring halo, and passes it to $\phi^v$
which predicts $\delta_i$ given the central halo properties.
 
\paragraph{Best-fit parameters.}
We list best-fit parameters for the discovered models in the paper in
\cref{tbl:params}.
The functional forms were extracted from the GN by approximating
both $\phi^e_1$ and $\phi^v$ over training data
with a symbolic regression and then 
analytically composing the expressions.
Although the symbolic regression fits
constants itself, this accumulates error from the two
levels of approximation
(graph net to data, symbolic regression to graph net).
Thus, we take out the functional
forms as given in \cref{tbl:params},
and refit the parameters directly to the training data.
This results in the parameters given, which
are used to calculate accuracy of the symbolic models.
 
\begin{table*}[]
\centering
\begin{tabular}{@{}lcccc@{}}
\toprule
& Test & Formula & Summed Component & ${\scriptstyle \ev{\abs{\delta_i - \hat{\delta}_i}}}$\\ \hline \hline
\multirow{2}{*}{\STAB{\rotatebox[origin=c]{90}{Old~}}} & Constant & $\hat{\delta}_i = C_1$ & N/A & 0.421
\\ \cline{2-5}
& Simple & $\hat{\delta}_i = C_1 + (C_2 + M_i C_3)e_i$ & $e_i = \sum_{j\neq i}^{\abs{\mathbf{r}_i - \mathbf{r}_j} < 20} M_j$ & 0.121 
\\ \hline
\multirow{2}{*}{\STAB{\rotatebox[origin=c]{90}{New~}}} & Best, without mass & $\hat{\delta}_i = C_1 + \frac{e_i}{C_2 + C_3 e_i \abs{\mathbf{v}_i}}$ & $e_i = \sum_{j\neq i} \frac{C_4 + \abs{\mathbf{v}_i - \mathbf{v}_j}}{C_5 + (C_6\abs{\mathbf{r_i} - \mathbf{r_j}})^{C_7}} $& 0.120 \\ \cline{2-5}
& Best, with mass & $\hat{\delta}_i = C_1 + \frac{e_i}{C_2 + C_3 M_i}$ & $e_i = \sum_{j\neq i} \frac{C_4 + M_j}{C_5 + (C_6 \abs{\mathbf{r}_i - \mathbf{r}_j})^{C_7}}$& 0.0882 \\
\bottomrule
\end{tabular}
\begin{tabular}{@{}lc@{}}
\toprule
\toprule
Test & Best-fit Parameters  \\ \midrule
Simple & $C_1= 0.415$  \\
Traditional & $C_1=-0.0376,  C_2=0.0529,  C_3=0.000927$   \\
Best, without mass & $\begin{array}{c}
C_1 = -0.199, C_2 = 1.31, C_3 = 0.027, \\C_4=1.54, C_5=50.165, C_6=18.94, C_7=13.21\end{array}$ \\
Best, with mass &   $\begin{array}{c}
C_1 = -0.156, C_2= 3.80, C_3=0.0809,\\
C_4 = 0.438, C_5 = 7.06, C_6 = 15.5, C_7=20.3
\end{array}$
\\
Best, with mass and cutoff$^\ast$ & $\begin{array}{c}
C_1 = -0.149, C_2 = 3.77, C_3=0.0789,\\
C_4 = 0.442, C_5 = 7.09, C_6 = 15.5, C_7=21.3
\end{array}$ \\
\bottomrule
\end{tabular}
\caption{Best-fit parameters for the functional forms 
used to estimate the overdensity of dark matter halos. The functional
forms are given in the upper table for reference.
$^\ast$Here we use the same formula
as ``Best, with mass,'' since we found an equivalent
formula by only looking at the 80\%
chunk of the data. The constants in that functional form
are also fit by only training on that fraction of the data.
}
\label{tbl:params}
\end{table*}